\documentclass[10pt,twocolumn,letterpaper]{article}

\usepackage{cvpr}              

\newcommand{\red}[1]{{\color{red}#1}}

\usepackage{pifont}

\definecolor{cvprblue}{rgb}{0.21,0.49,0.74}
\usepackage[pagebackref,breaklinks,colorlinks,allcolors=cvprblue]{hyperref}
\usepackage{multirow}
\usepackage{bm}
\usepackage[dvipsnames]{xcolor}

\newcommand{\green}{\textcolor{green}}

\usepackage{listings}
\def\paperID{*****} 
\def\confName{CVPR}
\def\confYear{2025}

\title{Robusto-1 Dataset: Comparing Humans and VLMs on real out-of-distribution Autonomous Driving VQA from Peru}

\author{Dunant Cusipuma$^1$, David Ortega$^1$, Victor Flores-Benites$^{1,2}$, Arturo Deza$^{1,2}$\\
$^1$Artificio \\ $^2$Universidad de Ingeneria y Tecnologia (UTEC)\\
Lima, Peru\\
{\tt\small \{dunant.c, david.ortega, vfloresb, deza\}@artificio.org}
}

\begin{document}
\maketitle

\begin{abstract}
As multimodal foundational models start being deployed experimentally in Self-Driving cars, a reasonable question we ask ourselves is how similar to humans do these systems respond in certain driving situations -- especially those that are out-of-distribution? To study this, we create the \href{https://huggingface.co/datasets/Artificio/robusto-1}{Robusto-1} dataset that uses dashcam video data from Peru, a country with one of the ``worst'' (aggressive) drivers in the world, a high traffic index, and a high ratio of bizarre to non–bizarre street objects likely never seen in training. 
In particular, to preliminarly test at a cognitive level how well Foundational Visual-Language Models (VLMs) compare to Humans in Driving, we move away from bounding boxes, segmentation maps, occupancy maps or trajectory estimation to multi-modal Visual Question Answering (VQA) comparing both humans and machines through a popular method in systems neuroscience known as Representational Similarity Analysis (RSA). Depending on the type of questions we ask and the answers these systems give, we will show in what cases do VLMs and Humans converge or diverge allowing us to probe on their cognitive alignment. We find that the degree of alignment varies significantly depending on the type of questions asked to each type of system (Humans vs VLMs), highlighting a gap in their alignment.

\end{abstract}    
\section{Introduction}
\label{sec:intro}

\begin{figure}[!t]
    \centering
    \includegraphics[width = \columnwidth]{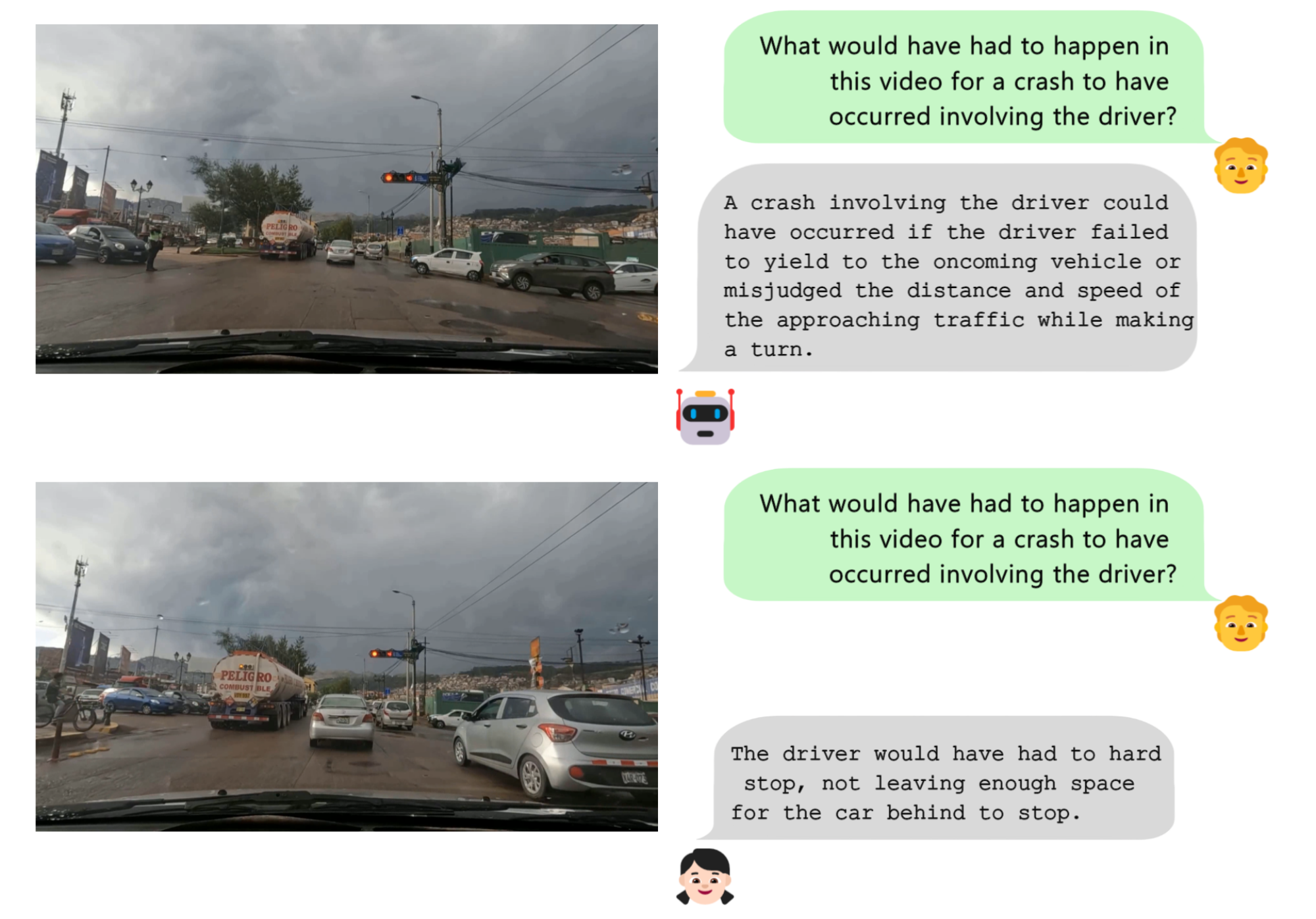}
    \caption{As multi-modal foundation models start being tested for Autonomous Driving applications, we inquire their cognitive alignment under a Visual Question Answering scheme of multiple videos comparing the answers of VLMs to those of Humans with tools from systems neuroscience. For this example in particular, a closer look reveals that the policeman is telling the driver to run through the red light. These sort of edge-case scenarios allow us to better probe cognitive 
 alignment.}
    \label{fig:GeneralSketch}
\end{figure}


\begin{figure*}[!t]
    \centering
    \includegraphics[width = \textwidth]{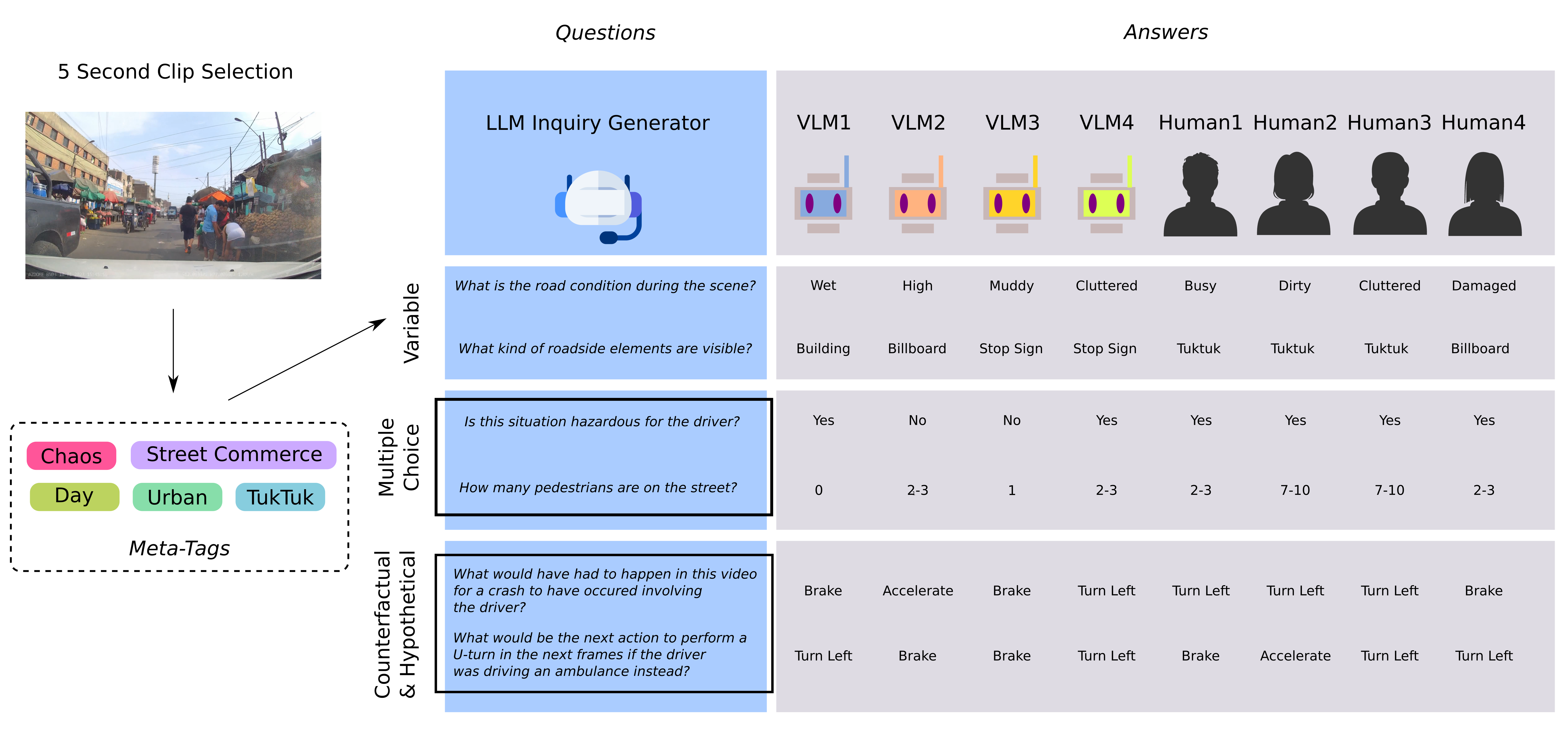}
    \caption{Overview of the VQA procedure on the Robusto-1 Dataset. A set of 5 second clips are seen by ground truth anotators (authors) and Meta-Tags are extracted from 16 different categories. These are then passed per each video to a ``Blind Oracle'' LLM that formulates a set of 5 variable questions per clip. An addition 5 set of multiple choice questions that have Yes/No answers and or involved rating or counting, and 5 open counterfactual questions are added to the total pool of 15 questions per clip. We then ask a group of VLMs and Humans these questions to collect their answers.}
    \label{fig:Annotation_Exp}
\end{figure*}

After several decades of development, self-driving cars are finally driving in San Francisco, London and Shanghai~\cite{waymoWaymoOpen}; and yet despite their initial success, accidents in the industry still occur due to miscalculation at different stages of the driving pipeline: be it at perception, navigation, decision making and/or control -- even in end-to-end trained systems~\cite{chougule2023comprehensive}. And while it is generally accepted that most of these errors occur at the decision making stage, it is currently unknown how well AI systems in Autonomous Vehicles (AV) are able generalize to current \textit{visual} out-of-distribution scenarios, and how the causal factor of crashes and accidents stem from such stages (perception, decision making, control). Indeed, at the perceptual stage, an active research questions is: how well will current AV systems perform when we push them to their limit?~\cite{han2021new}.


In industry more so than academia, a current alternative for out-of-distribution testing relies on simulated data \textit{(``GenAI'')} where prompt \& graphic engineers render a plethora of fake collisions or other OOD scenarios to train and test AV systems -- and while this is non-trivial, it is based mainly on anecdotal experience or biased creative input~\cite{ParallelDomain,AppliedIntuition}. We however find an interesting and realistic alternative: we collect data from Peru, a country with one of the ``worst'' (aggressive) drivers in the world, a high traffic index, and a high ratio of bizarre to non-bizarre street objects. Naturally, one must wonder how well will these AD systems generalize to these situations where reckless driving, tuk-tuks, street dogs and unpaved roads are the norm rather than the exception?


 
Indeed, it has not been the first time that pushing the limits of classical datasets in unconventional ways has previously been useful in computer vision and artificial intelligence~\cite{antol2015vqa,chandrasekaran2016we,baradad2021learning,harrington2022finding,chollet2024arc}. Recall that for classical object recognition, datasets such as Object-Net~\cite{barbu2019objectnet} and Stylized ImageNet~\cite{geirhos2018imagenet} have complemented the success of ImageNet~\cite{russakovsky2015imagenet}, teaching us about texture bias or cue-conflicting background information~\cite{NEURIPS2020_db5f9f42,dodge2017study,golpayegani2023clarifying}. Similarly, from a scope of domain generalization~\cite{wang2022generalizing}, we hope that the Robusto-1 dataset can be complementary to many other datasets in Autonomous Driving that focus on testing on clean driving conditions in USA, Canada, UK, Europe or China \& Japan, where the social norm is to \textit{follow the rules}, as is proper in first-world countries where self-driving cars are currently being deployed. However, if we can push the boundary in testing of what is generally not seen at training, then we can have a better assessment of how well these systems will perform when facing the unexpected. 


Thus our paper presents three main contributions: First, we will take advantage of a dataset that is gathered from Peru, that as argued previously is not only by nature \textit{out-of-distribution}, but it is also highly likely that our type of video data has never been seen in training~\cite{Ortega_Cusipuma_Flores_Deza_2023}; Second, we push the limit to comparing AV perceptual systems not through what is standard in terms of object recognition or semantic segmentation, but at the visuocognitive level through  Visual Question Answering (VQA). The importance of running these tests is critical given the explosion of multimodal foundational models that are generally openly available by big tech corporations and have started to be adapted for Autonomous Driving~\cite{marculingoqa}; Third, we use a Representational Similarity Analysis (RSA)~\cite{kriegeskorte2008representational,kornblith2019similarity} framework, that is originally from Systems Neuroscience and now pervasive in the neonascent field of NeuroAI~\cite{sucholutsky2023getting,Hosseini2024.12.26.629294,mineault2024neuroai}, to study the cognitive alignment between humans and VLMs in the context of (Autonomous) Driving.

Indeed, benchmarking multimodal foundation models on VQA is also useful because we have seen early experiments of self-driving cars driving with VLMs~\cite{tianlarge,renz2024carllava,marculingoqa}, signaling that this may be the way of the future. In addition, we hope that as the field of Autonomous Driving moves into releasing their foundation models, that this dataset and evaluation framework may be useful to compare their performance to humans, where data from every country in the world -- especially emerging economies -- given their \textit{out-of-distribution} nature, and unique driving ecosystem, can help Self-Driving cars become a reality worldwide.

\section{Dataset Construction}
\label{sec:Construction}

A total of 285 videos of varying length of at least 5 minutes each were recorded. From these videos, 200 five-second videos were sampled for the Robusto dataset. A total of 7 held-out videos were used for the preliminary VQA analysis in this paper. The initial set of 285 videos were collected through 2 cameras placed as dashcams from two sedan-like cars/vehicles. These cameras were one DASHCAM-Azdome-BN03 2k-1440p GPS and one GOPRO HERO 10 set to $1920\times1080$. These videos were collected from 3 cities in Peru: Lima, Cusco and Cajamara. Peru was chosen as a first iteration for the Robusto dataset because it has the 2nd worst drivers in the world, the highest traffic index of Latin America, and the 4th highest rank in Global poorest quality of road infrastructure in 2023 according to studies by \textit{Compare The Market}~\cite{Hadnagy_2023}, \& \textit{\href{https://www.tomtom.com/traffic-index/ranking/}{TomTom's Traffic Index}}.

Inspired by the Visual-Language Model evaluations of LingoQA~\cite{marculingoqa}, we then selected 200 5-second scenes from the database sampled at 10 \textit{Hz}. These \textit{50} frames per video/scene contained a mixture of interesting events ranging from jaywalking, reckless driving in varying levels of clutter from which we then later formulate a set of questions for VQA (see Ortega~\textit{et al.}~\cite{Ortega_Cusipuma_Flores_Deza_2023} on why this is useful).






\subsection{Question Formulation}

For each 50 frame clip we select a total of 15 questions per clip to show to both humans and machines. These 15 questions are composed in 3 main groups: 1) Variable questions, that are exclusive to the image and that were prompted by an Oracle LLM that has access to the annotated Meta-data~\cite{berrios2023towards} -- these have an open-ended answer but are mostly factual and have a more objective notion of ground truth (See Table~\ref{table:Questions1to15}); 2) Multiple choice questions that were formulated by the authors of the paper covering a wide range of topics and that have a Yes/No answer-like component or that will require a rating/counting-like answer; 3) Counterfactual \& hypothetical questions that are also formulated by the authors and will be shared for every single clip and will require advanced reasoning \textit{(``what-if X or Y happens?'')} from humans \& VLMs, and have an open-ended answer with no objective ground truth.

\subsubsection{Block 1: Variable Questions}
As mentioned earlier, the variable questions pertain to the specific video in question. These questions are based on metadata, which is why they are variable per video, as the environments depicted in the videos differ. The questions may include: \textit{``What is the driving maneuver performed in this scene?''}, \textit{``What is the traffic condition in this scene?''}, \textit{``What is the weather condition?''}, \textit{``Is there any pedestrian activity?''}, etc. The significance of these questions lies primarily in the responses provided by the Oracle LLM, and most of these questions have a high degree of objectivity.

\subsubsection{Block 2: Multiple Choice Questions}
\begin{itemize}
\item \textbf{Q6: ``Please rate the level of clutter from 1 to 10; 10 being the highest level of clutter and 1 being the lowest.''} This question is of high relevance because it will require spatio-temporal reasoning of the 5 second clip in addition to a high level perceptual task such as clutter, which is somewhat ambigious despite having image-computable models that still shown some variation~\cite{rosenholtz2007measuring,yu2014modeling,deza2016can,nagle2020predicting,kyle2023characterising}. 
\item \textbf{Q7: ``Is this a recurrent driving scenario  for you?''} This Yes/No question is quite interesting for humans because we can assess how often it is that they encounter certain out-of-distribution events given the nature of each clip. Moreover, this question is interesting for machines because it probes a sense of meta-awareness as the VLM that has never really driven a car~\cite{thrush2024strange,tuckute2024language}, yet forcing it to give a yes/no answer may give us insights of the training data it has been fed with.
\item\textbf{Q8: ``How many pedestrians are there in the scene?''} Both humans and VLMs will likely have a hard time counting pedestrians in a moving scene, so this question is important especially in crowded conditions where it is likely both humans and VLMs will have a hard time likely providing ball-park estimates. It is important to also notice that VLMs are generally not good counters~\cite{luo2024vision,ma2024large,thrush2024strange,kosoy2025decoupling}, so having an open question like where they must integrate identity spatio-temporally will be important even if prompted with an ordinal range of intervals (0,1,2-3,4-6,7-10,11-20,21+).
\item\textbf{Q9: ``Is this situation hazardous for the driver?''} 
Assessment of danger with a Yes/No response is an important cue for driving in both humans and VLMs thus adding this constraint is highly important for alignment \& morality research in the future of autonomous vehicles~\cite{de2020doubting,de2021deliberately,lecun2022path,sucholutsky2023getting}
\item\textbf{Q10: ``On a scale of 1-10, how well do you think an autonomous vehicle would perform in this scene?''} This last rating question tackles two interesting sub-questions which are how dangerous the scene is, and a theory-of-mind angle of transplanting the human or the VLM in an Autonomous Vehicle. This of course is interesting because the assumption is that humans will assess greater levels of danger in terms of assessing a difficult situation compared to a machine; while such VLM's may possibly underscore the difficulty of navigating a scenario. 
\end{itemize}
\begin{figure*}[!t]
    \centering
    \includegraphics[width = \textwidth]{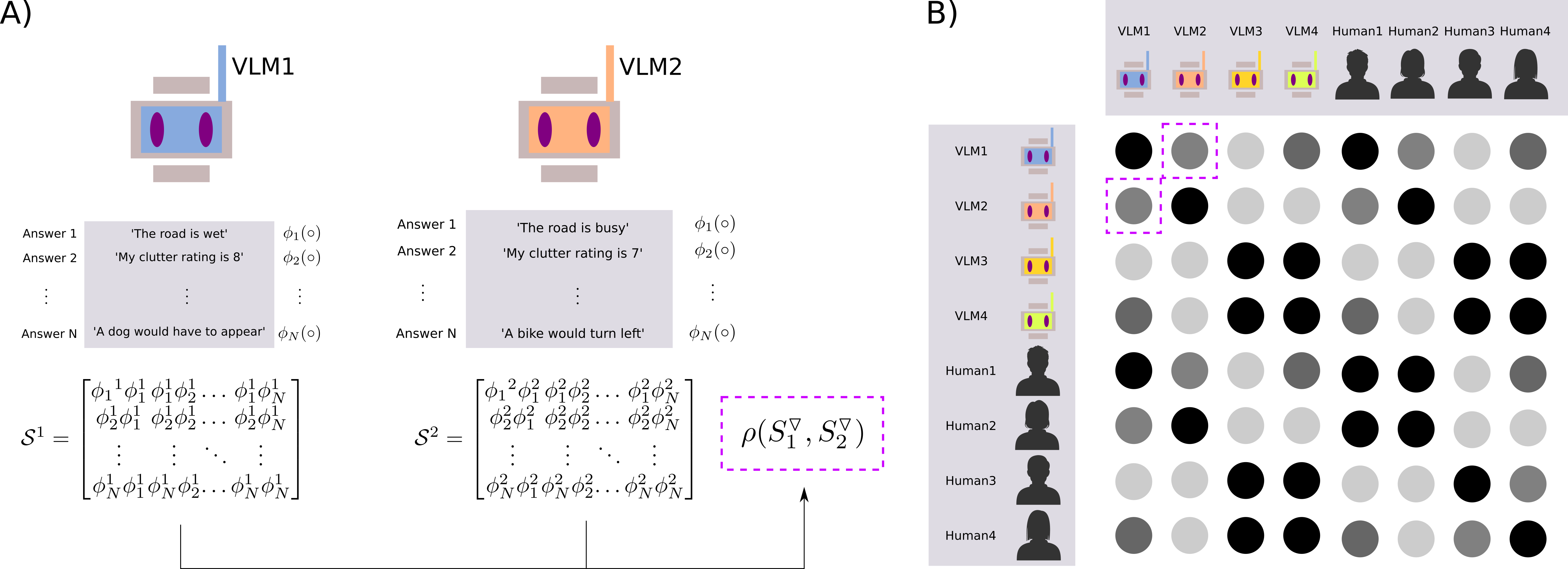}
    \caption{A figure that shows how to calculate the System Similarity Matrix through Model Gramians as done in Representational Similarity Analysis (RSA)~\cite{kriegeskorte2008representational}. A) We transform each answer into a vector through an embedding to later calculate each system's Gramian. Upper triangular parts of the Gramians across two systems are then correlated (violet). This can be applied to both humans and VLMs. B) The system similarity matrix $\mathcal{M}$ calculated over all humans and machines allows us to get an idea of how each system is similar to one another. A cartoon with no real values is shown in this diagram.}
    \label{fig:Gramian}
\end{figure*}

\subsubsection{Block 3: Counterfactual \& hypothetical questions}

We finally also decided to add a set of 5 constant counterfactual \& hypothetical questions to each human and machine in our dataset. A counterfactual is known as a type of question similar to a hypothetical ``what-if?'' scenario except that ~\cite{gerstenberg2024counterfactual} it is based on the knowledge of the outcome. Mainly, hypotheticals are of the type: \textit{``what would you do if?''} (focus on the future), while counterfactuals are of the type: \textit{``what would you have done if?''} (conditioned on the knowledge of the real outcome). Both counterfactual and hypothetical research has been shown to test interesting cognitive biases of each system that will likely arise due to differences in training data, learning algorithm, architecture, inference procedure, and in the case of humans, potentially cultural biases~\cite{lagnado2013causal,zanardi2023counterfactual,hsu2023interpretable} -- and only recently have they started to be explored with rigor in Autonomous Driving~\cite{de2021driverless,kirfel2023anticipating,tsirtsis2024towards}. We list three counterfactual questions and two hypothetical questions that will be asked for every short 5 second clip to each human and machine. These are:

\begin{itemize}
    \item \textbf{Q11: What would have had to happen in this video for a crash to have occured involving the driver?}
    \item \textbf{Q12: What would have had to happen in this video for an external crash to have occured not involving the driver?}
    \item \textbf{Q13: What would have happened in the scene if you had done the opposite action (brake vs accelerate, or accelerate vs brake) ?}   
    \item \textbf{Q14: What would be the next action to perform a U-turn in the next frames if the driver was driving an ambulance instead?}
    \item \textbf{Q15: What would be the next action to perform a U-turn in next frames if the driver was driving a motorcycle instead?}
\end{itemize}

\section{Comparing Humans and VLMs through Systems Neuroscience}
\label{sec:Analysis}

In systems neuroscience, representational similarity analysis (RSA) was created to properly study the degree of similarity between any two biological or artificial systems~\cite{kriegeskorte2008representational,yamins2014performance,khaligh2014deep}. However, the main problem when trying to compare an artificial neural network to a biological system such as a monkey's brain, is that often times the dimensionality of the feature vector responses of both systems are not compatible with each other.

Let us consider the following: suppose a feature vector $(f_A(\circ))$ for the hidden layer of a neural network has dimensionality $d_{A}>1000$ (5th layer of AlexNet), while the biological sensor array $(f_B(\circ))$ has dimensionality $d_{B}<100$ (for example the average spiking rate of a neural array that can record from 80 neurons), how can we compute a correlation or notion of similarity between both systems when the dimensionality of the space that we are sampling from is \textit{different}, even though they are exposed to the \textit{same} input stimuli like an image $I$? 

Given the previous limitation, we can not compute a well-known error function such as Mean Square Error (MSE) across the two observations because the dimensions of such observations do not match, i.e. $(f_A(I))-(f_B(I))$ is noncomputable. The solution to this dimensionality incompatibility $(d_A\neq d_B)$ as posed by RSA is that we feed both systems the same $N$ images and create an $N\times N$ Gramian Matrix that normalizes the dimensionality by creating a matrix of inner products of the feature responses to themselves across a wide variety of such input images. In that sense we are implicitly studying the topography of each system and their convergence/divergence with respect to each other~\cite{mineault2024neuroai}.

Luckily, we are in a very fortunate situation where we do not have a dimensionality incompatibility issue for our analysis since both Humans and VLMs will respond to the same questions, and, although there can be differences in terms of varying length of answers per question across all systems, these are normalized to the same dimensionality through a sentence embedding~\cite{reimers2019sentence,song2020mpnet}. This implies that we~\textit{can} calculate other measures of answer variability such as the euclidean distance between any two answers or one answer and the median of all answers per question  -- a more meaningful metric -- as we will see later in the Experimental section of this paper. That being said, applying Representational Similarity Analysis when the dimensionality is preserved across all systems is just as applicable to us -- even more so because knowing ground truth is \textit{irrelevant} in our experiments~\textit{(!)}. 

Recall that to make such comparisons, even humans themselves will disagree with one another given the ambiguity of certain questions and the stimuli shown where in many cases there is no objective ground truth answer (\textit{e.g.} the counterfactual and hypothetical questions). We will discuss more about why knowing ground truth is irrelevant in our experiments in the next sub-section~(Sec~\ref{sec:RSA}) by expanding more on the methodology behind RSA, though a more detailed review can be seen in Kriegeskorte~\textit{et~al.}~\cite{kriegeskorte2008representational}.

\subsection{Representational Similarity Analysis}
\label{sec:RSA}

To calculate the Gramian Matrix of each System we must first transform each answer (a sentence) to a vector through an embedding $\phi(\circ)$, the characteristic Gramian Matrix of each system $\mathcal{S}$ given the contexts of responses is:

\[
\mathcal{S} = 
\begin{bmatrix}
\phi_1\phi_1 & \phi_1\phi_2 & \dots & \phi_1\phi_N \\
\phi_2\phi_1 & \phi_2\phi_2& \dots & \phi_2\phi_N \\
\vdots & \vdots & \ddots & \vdots\\
\phi_N\phi_1 & \phi_N\phi_2 & \dots & \phi_N\phi_N
\end{bmatrix}
\]

Taking inspiration from Representational Similarity Analysis (RSA) in Computational Neuroscience~\cite{kriegeskorte2008representational}, we create these System Gramians for each model that is representative of the answers each model has given to the set of $N=105$ questions (7 videos $\times$ 15 questions). The goal of this approach is to then summarize the similarity across all systems by creating a correlation matrix by pitting each System's Gramian $(\mathcal{S})$ against each other by only taking the upper triangular component $(^\triangledown)$ of such Gramians, and correlating them through the pearson correlation $(\rho)$. This yields the following cross-systems summary matrix $\mathcal{M}$: 
\[
\mathcal{M} = 
\begin{bmatrix}
\rho(S_1^\triangledown,S_1^\triangledown) & \rho(S_1^\triangledown\,S_2^\triangledown) & \dots & \rho(S_1^\triangledown,S_M^\triangledown) \\
\rho(S_2^\triangledown,S_1^\triangledown) & \rho(S_2^\triangledown,S_2^\triangledown) & \dots & \rho(S_2^\triangledown,S_M^\triangledown) \\
\vdots & \vdots & \ddots & \vdots\\
\rho(S_M^\triangledown,S_1^\triangledown) & \rho(S_M^\triangledown,S_2^\triangledown) & \dots & \rho(S_M^\triangledown,S_M^\triangledown)
\end{bmatrix}
\]

Notice that $\mathcal{M}$ is also symmetric, of size $M$ (total number of humans and GenAI systems), and has all its diagonal elements set to $1$ given the autocorrelation of each system to itself; furthermore all elements are bounded in the interval $\big[-1,1\big]$.

\subsection{Metric-based Analysis}
\label{sec:MetricL2}

An additional type of analysis we have the option to run given equality of dimensions of all embeddings, is \textit{distance-based} from the embeddings $\phi(\circ)$ in their latest space. In particular to assess similarity of all systems to each systems, we will compute and visualize the L2 distance to the median of all responses per question per video for all systems. This allows us to find a reference point (the median) from which we can compute a distance excluding outliers to know how similar systems are depending on the distance they have to such reference point from a uni-dimensional perpsective. The L2 distance of a system to the median per question $i$ and video $j$ is determined by:

\begin{equation*}
    \text{L2}(\phi^A_{i,j}(I),Median( \phi^{\circ}_{i,h}(I)))
\end{equation*}

\subsection{Dimensionality Reduction}
\label{sec:dimRed}

We will then finally view the results in the simplest form of analysis by performing PCA to the first two main principal components divided of all the systems' embedding vectors divided by block (1,2, and 3). This qualitative description should allow us to visualize the raw overlap of answers across all systems (humans and VLMs) before following-up with quantitative analysis such as the Distance to the Median and RSA. 



\section{Experiments}
\label{sec:Experiments}

\begin{figure*}[!t]
    \centering
    \includegraphics[width = \textwidth ]{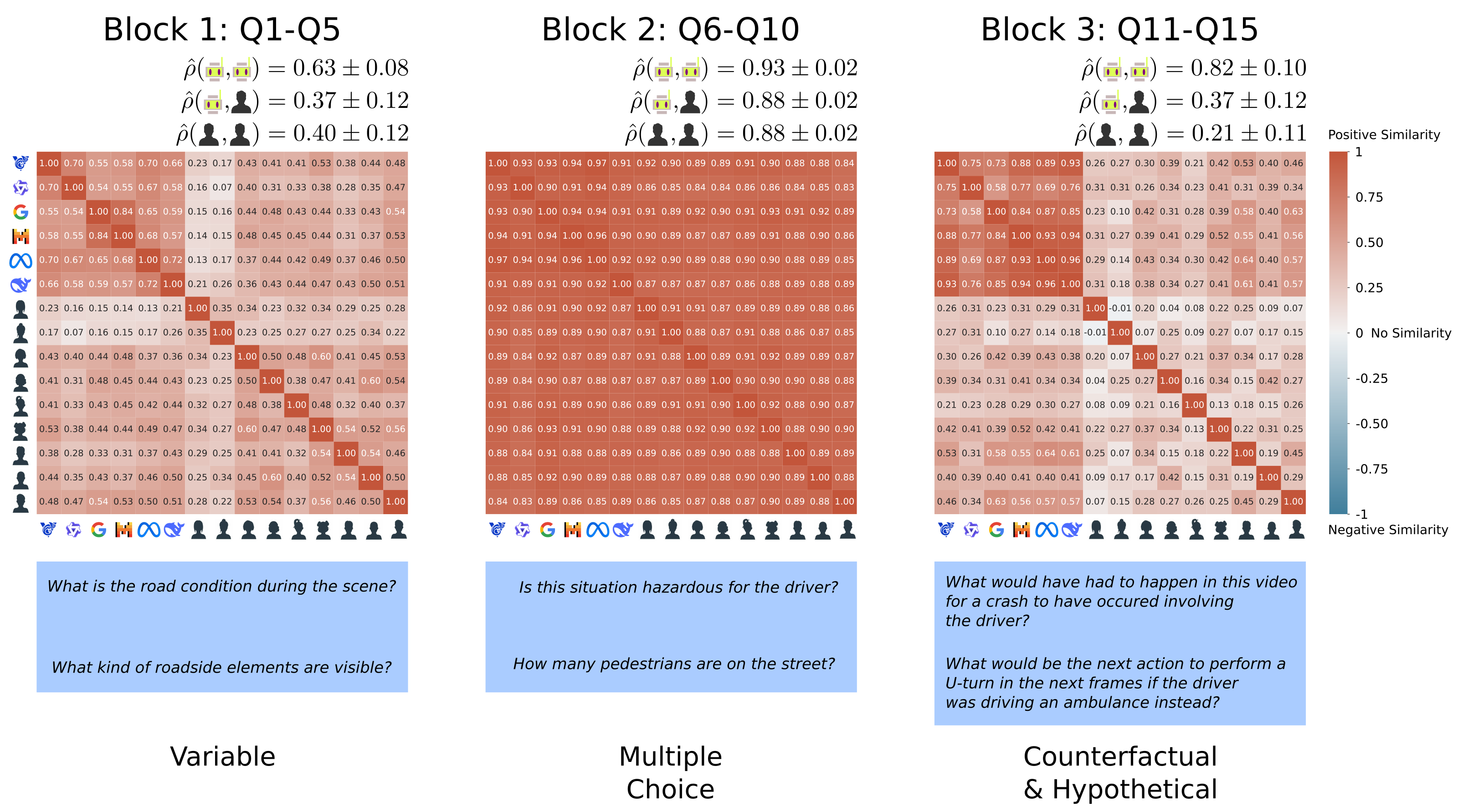}
    \caption{The first general result we find after applying Representational Similarity Analysis (RSA) to responses of both humans and VLMs, is that system convergence and divergence is modulated by the type of questions asked. Broadly speaking, we find that all VLMs respond very similar to each other independent of the types of questions asked with a surprisingly high correlation for counterfactuals \& hypotheticals. Humans on the other hand diverge heavily for counterfactual \& hypotheticals and converge strongly for multiple-choice.}
    \label{fig:Results_RSA}
\end{figure*}


The main experiment of this paper had 9 volunteer Human subjects who gave their digital consent for public use of their anonymized data, and 6 VLM models what were shown the same subset of 7 videos each with 15 questions per video. In what follows in the paper, we will analyze and draw interim conclusions of the convergence and divergence of each system both across and between each group (humans vs VLMs, humans vs humans and VLMs vs VLMs). While we did not set out to prove or disprove any initial hypothesis, we were inclined to predict that most VLMs would respond very differently to each other given the different \textit{``ethos''} that each company has regarding compute and training algorithms; and that conversely Humans would present a high degree of similarity in their responses. To our surprise, we found that in some cases the results seemed to be the opposite, with exceptions of high partial overlap across all systems for the multiple choice questions.

\subsection{Representational Similarity Analysis}
\label{sec:RSA_Results}

We first begin by analyzing the results from computing the similarity matrices through the Gramian matrices of each system as done in RSA (Section~\ref{sec:RSA}), and separate these analysis in 3 blocks. Block 1 consists of the first questions 5 questions (1 to 5) that are composed of variable video-specific questions; Block 2 consists of 5 multiple-choice questions (questions 6-10), and Block 3 consists of questions 5 counterfactual \& hypothetical (questions 11-15). These are shown in Figure~\ref{fig:Results_RSA}.

Broadly speaking, we find that VLMs all share a high rate of similarity in their answers independent of the type of questions asked. Humans on the other side, have a greater degree of variance. With a very high level of alignment in the multiple choice questions, then the variable ones, and finally almost no similarity in the counterfactual \& hypothetical questions. When looking at Figure~\ref{fig:Results_RSA}, Block 2 shows a relativaly high degree of similarity across all systems (Humans and VLMS). This in a way is expected because the degrees of freedom for the answers in the multiple choice questions is quite low as there are some questions that are Yes/No, and others that have 10 options (for example the 1-10 clutter rating question). Table~\ref{table:Questions1to15} in the supplementary material summarizes these types of questions.

\begin{figure*}[!t]
    \centering
    \includegraphics[width = \textwidth ]{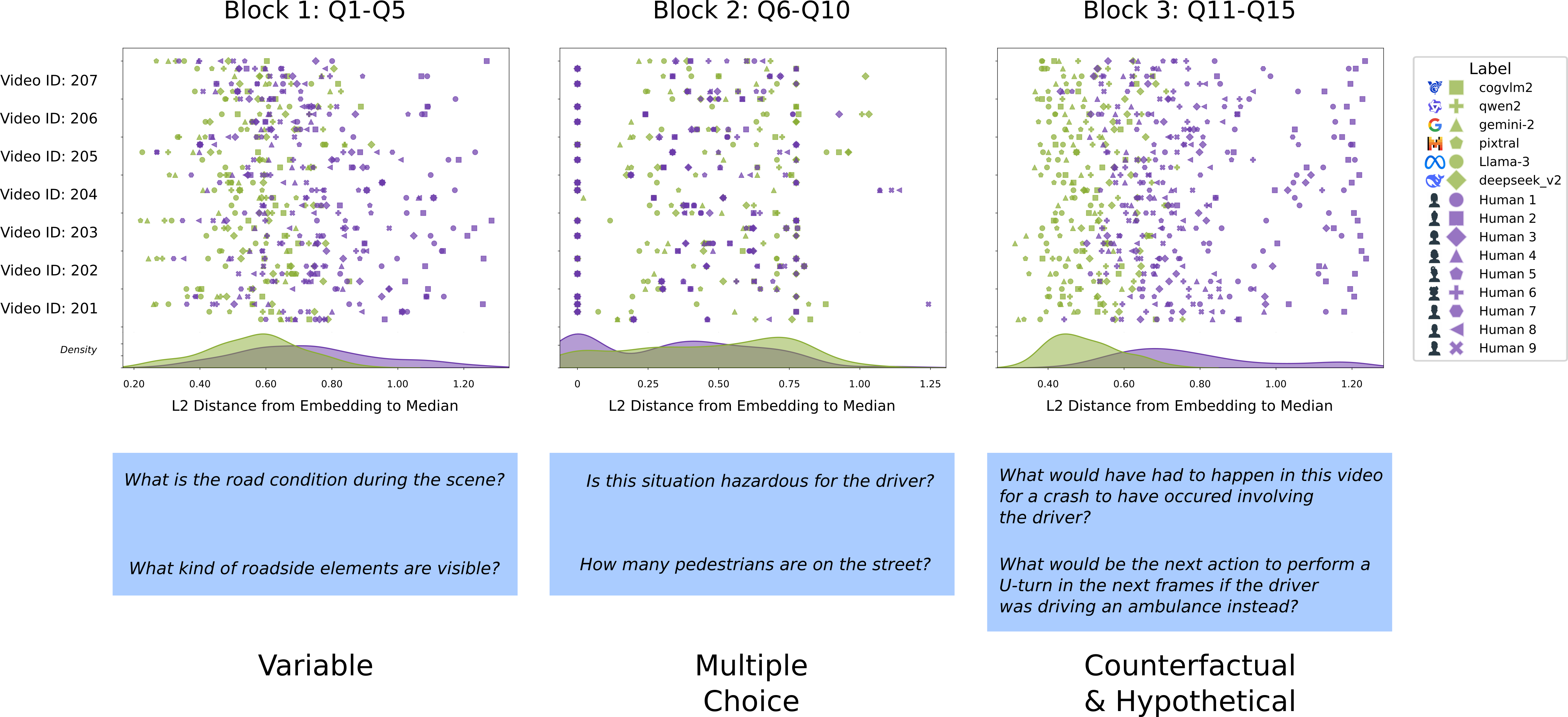}
    \caption{In this figure we show the distance of each response per question across all systems to the median response. Responses placed here for the VLM was the average response per question rather than a single response. We generally observe that the overlap across the answers for VLMs and Humans shifts depending on the nature of the questions asked with a larger partial overlap for block 2 given the nature of multiple- choice questions and the smaller space that answers can space as they are prefixed. Variance for block 3 on the other hand is larger across humans and VLMs given the complexity of counterfactual \& hypothetical questions.}
    \label{fig:Results2}
\end{figure*}


\subsection{Inter-System Agreement}
\label{sec:Inter_Results}

To better understand at a more fine-grained level how each system's answer geometrically steered the overall system's topology for the similarity matrices, we proceeded to perform an inter-system agreement analysis by computing the distances between each embedding to the median response per answer across all systems for every question and answer pair. This allows us to probe how \textit{aligned} Humans and VLMs are to each other both between and across systems at a per question basis (Humans vs Humans, VLMs vs Machines, and VLMs vs VLMs). The results from this analysis can be shown here in Figure~\ref{fig:Results2}.

We find once again that there is a difference in the distribution of the pattern of answers depending on the nature of the questions asked to both humans and VLMs. In some cases the answers are more similar (see for example Block 1) likely due to the natural of the questions -- although they are open they are less ambiguous. Similarly, block 2 despite being multiple choice shows that the raw pattern of answers are very different across both humans and VLMs in the two bi-modal distributions have peaks placed on opposite ends of the spectrum -- highlighting greater biases, although there seems to be one mode of the bi-modal distribution that is shared. Finally block 3 shows the largest difference across Humans and VLMs, likely due to the complex nature of the questions (these are counter-factual and hypotheticals). Humans and VLMs answer these last group of questions very differently, and while we can not assume that the way they answer is different from the way they think (internal representations vs behavioural answers), it does shed light on such cognitive differences. 


\subsection{Dimensionality Reduction}
\label{sec:DimRed_Results}

Finally, to have a global assessment of how the general pattern of raw responses get mapped into a 2D plane across all systems, we projected all answers depending on what block they belong to from each system into a 2D plane with PCA, respectively yielding a $22\%$, $52\%$, $29\%$ explained variance. These plots show that there is a partial overlap across both systems, but such results are misleading as verified by the explained variances , and because some plots (Block 3) seem to suggest a high alignment between humans and machines, but this is not true until looking into our previous analysis from Sections~\ref{sec:RSA_Results},\ref{sec:Inter_Results}, highlighting the importance of RSA.





\begin{figure*}[!t]
    \centering
    \includegraphics[width = \textwidth ]{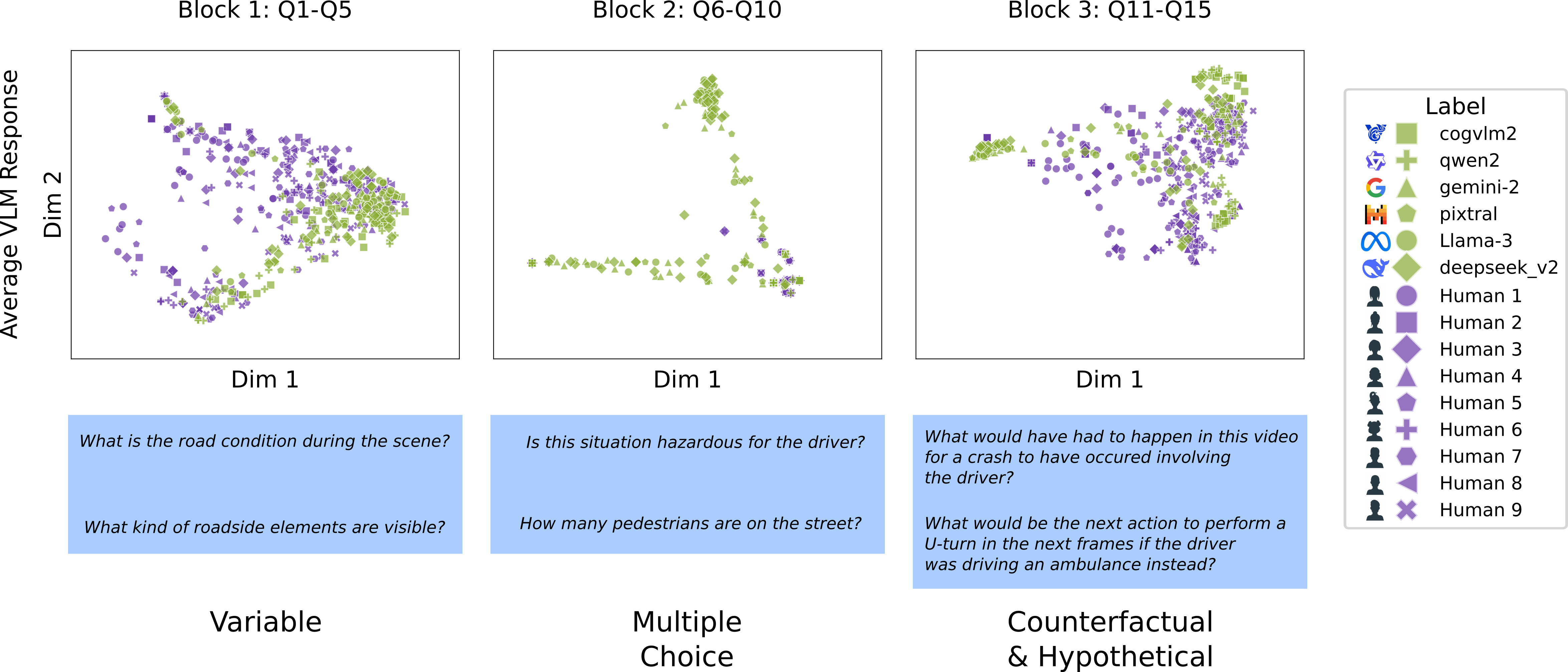}
    \caption{A collection of 2D projections of the answers from both Humans (violet) and VLMs (green) divided by Blocks. We notice that Humans and VLMs have partial overlap but they generally answer questions very differently, and these vary depending on the type of questions asked. These average embeddings per answer also highlight the variability of answers VLMs have to the same questions.}
    \label{fig:Results_Embed}
\end{figure*}
\section{Discussion}
\label{sec:Discussion}

In this paper we have barely scratched the surface through a preliminary study with humans and VLMs to know if AI-Chatbots can drive, and more precisely -- if they would react and behave in a similar way to us through Visual Question Answering (VQA). Of course this broader question requires more in-depth research not only at the autonomous driving level by doing trajectory prediction in a simulated 3D or real world environment, but also at a behavioral level through Q\&A experiments~\cite{marculingoqa}, and by internally probing representations in Humans and Machines (for now VLMs, commonly known as AI-Chatbots with multi-modal capacity). And while it is difficult to answer this question, in this discussion section we hope to fill in the pieces of this puzzle through understanding the results of some of our experiments, allowing us to build a scaffolding into what future experiments and research in this direction could look like.

First it is important to decouple our results -- while interesting -- with the fact that for any two systems that have similar language responses that have been inferred from an experiment \textit{does not} imply that the systems think similarly, although there could be a \textit{hidden correlation}. This is because, for two systems to be representationally aligned, system responses to multiple stimuli (be it images or text), they should share a common response at the behavioural stage, but also throughout the internal feature space of a system~\cite{sucholutsky2023getting,mineault2024neuroai}. In mathematical terms, we believe it is ``necessary, but not sufficient'', because two systems could have articulated the same answer, independent of whether they are both representationally aligned or not. This has been recently discussed in the \textit{NeuroAI Turing Test}~\cite{feather2025brain}. It is of course also possible (though less likely), that two systems have very similar internal representations but their answers diverge due to later-stage processing differences in the internal model of each system. This is why benchmarking using human data in a behavioral psychophysical setting has been a good first stepping stone for research in the LLM space through platforms like \href{https://lmarena.ai/}{LMArena} or \href{https://huggingface.co/spaces/open-llm-leaderboard/open_llm_leaderboard#/}{OpenLLM}, but still requires testing the alignment of their internal representations. 

All of this is to say that perhaps the future of research towards building safer Autonomous Driving systems would imply comparing not only outputs from humans answering surveys (as we did in this paper and where most VQA research is headed), but also advanced behaviour \& internal representations (\textit{e.g.} comparing the eye-movements or brain activations of an experienced NYC Taxi Driver in a MRI of MEG scanner with the attention maps and feature activation maps of a VLM) -- and while these sound like  futuristic experiments, current methods in computational neuroscience already make this possible~\cite{schrimpf2018brain,tuckute2024language,schrimpf2021neural,subramaniam2024revealing,harrington2022finding}.

Zooming out, we have found that there is
some alignment of VLMs to Humans but this is quite mild compared to how aligned VLMs are aligned to each other, or how divergent Human responses are to each other given the contrived questions they were prompted with. Why do most VLMs behave so similarly as we have inferred given their answering patterns? We believe it has to do with the curse of training with the same large pool of data: \textit{The Internet} and most models being based on a variant of a Transformer~\cite{vaswani2017attention}. As VLM models scale up due to commercial hype, and the ``data is all you need'' hypothesis (\textit{aka} The Bitter Lesson~\cite{sutton2019bitter}), they still fail to align with humans in how they answer questions in critical case scenarios like driving\footnote{Of course, this is a debatable goal because Humans are imperfect drivers and are likely not the long-term gold standard, but VLMs have not achieved super-human prowess~\cite{schrittwieser2020mastering} in general VQA~\cite{de2023visual,ming2024faitheval,liu2024mmbench,duan2024vlmevalkit,huang2024large,ku2025theoremexplainagent} nor driving VQA~\cite{marculingoqa,rekanar2024optimizing}, to allow us to make a claim that it is \textit{an advantage} that all VLMs respond the same way and \textit{different} than humans.}.

Indeed, it seems that it does not matter where a VLM is trained, what data was used, or what architecture was assigned, they were all highly similar. We would have liked to see an outlier VLM in our study that had a stronger difference to the others, in addition to similarity to a particular human or group of them. However, we also do see that even across humans, many of them diverge strongly when answering such questions shedding light on the complexity of the representational alignment problem~\cite{kriegeskorte2008representational}. Following the Anna Karenina principle of \textit{``All happy families are alike; each unhappy family is unhappy in its own way.''}, we observe something similar for VLMs and Humans in the driving context for this preliminary experiment where \textit{``All VLMs are alike, each Human is different in their own way''}. And while this may not be a negative outcome, it does shed light on the fact that most VLMs are not as different as we think they are despite a continuous race towards building intelligent machines that may one day drive on their own~\cite{huh2024platonic,hosseini2024universality}. 
\pagebreak
\newpage

\section{Acknowledgements}
We'd like to thank Tobias Gertenberg, Greta Tuckute, Robert Geirhos for insightful discussions over email, and Stanford's Wu Tsai Neuroscience Institute for feedback. This work was privately funded by Artificio and UTEC. All experimental surveys for data collections were approved by the Artificio Co-Founding team who are also authors of this paper. VLMs were ran locally, or through third party APIs, and through the GPU resources of the Google Cloud Start-Up Program. 

All human subjects (Peruvian citizens between the ages of 18 and 30) digitally consented to the use, release and publication of this data with proper anonymization as a preliminary study for research purposes in the field of Autonomous Driving and Artificial Intelligence.

All Human and VLM raw data \& code is open-sourced in the following HuggingFace repository: \href{https://huggingface.co/datasets/Artificio/robusto-1}{Robusto-1}

All work was done in Lima, Peru. All authors contributed significantly in the design, construction \& analysis of the experiments in addition to the writing of this paper. 

\newpage
{
    \small
    \bibliographystyle{ieeenat_fullname}
    \bibliography{main}
}

\clearpage
\setcounter{page}{1}
\maketitlesupplementary


\subsection{Human Protocol VQA}

A total of nine humans participated in this small pilot experiment as \textit{volunteers}. A consent digital consent form was given to the volunteers where they were briefly told about the goals of the study. Participants were required to perform the task on a computer or laptop and were not allowed to use their phones to ensure a wider field of view, as watching a video on a phone may result in missing key elements in such short clips. Responses were recorded digitally and stored anonymously with encrypted participant IDs. Participants provided their digital consent by ticking on a box in a Google Forms spreadsheet to share their data in anonymized way for research and commercial purposes. 

The participant demographics consisted of nine individuals aged 18 to 35 from Peruvians living in Peru. Subjects were recruited as a mixture of friends and colleagues of the authors through open advertising in different group chats. The participants had varying levels of driving experience and were fluent in English, as the questions were asked and answered in English. Participants also digitally confirmed their english fluency, and participants who did not have such were potentially going to be removed from the analysis. This was not the case and we analyzed all 9 subjects in the experiment. It is important to note that the VLMs were also tested using the same questions in English. All participants were Peruvian. We are aware that the small study group is interesting because Peruvians generally speak spanish (not english), and that VLMs have likely not seen dashcam driving data in Peru (a spanish speaking country). A future study will include English-speaking participants 
(\textit{e.g.} Americans) and show them a mixture of data from both people driving in the United States and Peru to study the interaction of language fluency and dashcam data provenance to the study.

Interestingly, some participants at the end of the experiment thought it was a text-base labelling task (given what they have read in the news about manual bounding-box labeling being required to train AI models), as they were unfamiliar with Question-Answering (QA) research. Approximately half of the participants reported via email, Slack or WhatsApp that many questions seemed subjective -- however, this is not a negative comment, as it verifies the intention of our experiment to push the boundary of human interpretability through OOD stimuli with questions of varying level of subjectiveness such as the hypotheticals \& counterfactuals in Block 3.

\subsection{Multimodal Input Processing Pipeline}

To systematically evaluate the ability of Vision-Language Models (VLMs) to analyze driving scenes, we implemented a input processing protocol tailored to the specific requirements of each model. The objective was to ensure that all models received equivalent multimodal input while respecting their individual API constraints and format requirements. This protocol enabled us to compare their performance fairly across tasks involving video-based visual question answering (VQA).

Each model was provided with a series of frames extracted from driving videos alongside a set of structured questions. Given the variability in how different VLMs process visual inputs, we employed a prompt adaptation mechanism that converts video data into a compatible format for each model. 

Below, we describe the input processing strategy for each VLM tested in our experiments.

\paragraph{CogVLM.}
For CogVLM, video data were submitted as complete binary files. The input video was read from a local file in binary mode and passed along with an adapted prompt via the Replicate API. The input dictionary included keys for the prompt, the binary video file (``input\_video''), and generation parameters (``top\_p'' set to 0.9, ``temperature'' set to 1, and ``max\_new\_tokens'' of 2000).

\paragraph{Qwen 2.}
For Qwen2, videos were hosted remotely. Each video was downloaded using HTTP requests, converted into an in-memory file using Python’s \texttt{BytesIO} module, and then combined with the adapted prompt to form the input. These data were sent to the model via the Replicate API in a similar structure as for CogVLM, enabling Qwen2 to process video inputs directly from remote sources.

\paragraph{Pixtral.}
Pixtral Large model processes video content by analyzing individual frames rather than receiving a complete video file as a single input.
For each video, frames were extracted at a rate of 1 FPS and converted to Base64-encoded JPEG strings. The resulting input was constructed as a message comprising a text component (the adapted prompt) followed by a series of image components. Each image was represented by a Base64 string prefixed with \texttt{"data:image/jpeg;base64,"}.

\paragraph{DeepSeekV3.}
DeepSeek V3 was evaluated by extracting video frames at 10 FPS and converting each frame into a Base64-encoded string. The adapted prompt was combined with a list of these image strings (each prefixed with \texttt{"data:image/jpeg;base64,"}) into a message structure, which was then submitted to the model via its API. 

\paragraph{Gemini.}
Gemini processes video inputs by first combining the system prompt with a marker that denotes the start of the visual sequence. Each video frame is then converted into an image component via the API’s \texttt{Part.from\_image} method, and a text component containing the user prompt is appended. Gemini 2.0 was deployed on Google Cloud Platform (GCP) through Vertex AI, utilizing the checkpoint “gemini-2.0-flash-exp” in accordance with the guidelines provided in the Vertex AI Generative Models 
\href{https://cloud.google.com/vertex-ai/generative-ai/docs/reference/python/latest/vertexai.generative_models}{documentation}. 
The LLM generation parameters were set to a maximum of 100 tokens, a temperature of 1.0, and a top-p of 0.9. 

\paragraph{Llama.}
For Llama-based models, our protocol transforms each video frame into a Base64-encoded JPEG string that is then integrated with the system instructions and user prompt into a single text block. This combined input is submitted to the model via its API. In our experiments, Llama 3.2 was deployed in GCP using Vertex AI, employing the checkpoint “Llama-3.2-11B-Vision-Instruct-meta.”
The LLM generation parameters were set to a maximum of 100 tokens, a temperature of 1.0, and a top-p of 0.9.

\subsection{MetaData \& Tags}
The distribution of driving scenarios suggested that we create a pre-fixed list of 16 meta-tags from which we manually annotate certain properties from a video clip. Sample meta-data attributes are: \textit{1. Vehicle Action}, \textit{2. Driving Action Reasoning}, \textit{3. Vehicle Motion Behavior}, \textit{4. Traffic Signs}, \textit{5.Traffic Lights}, \textit{6. Weather Conditions}, \textit{7. Road Surface Conditions}, \textit{8. Road Structures}, \textit{9. Static Objects}, \textit{10. Other Vehicle Behaviors}, \textit{11. Pedestrian Behaviour}, \textit{12. Unexpected Obstacles}, \textit{13. Emergency Situations}, \textit{14. Lighting Conditions}, \textit{15. Traffic Conditions}, \textit{16. Driving Environment}.
The full list of information of the labels derived from the meta-data attributes can be seen in the Table~\ref{table:MetaTags}. These meta-tags are available for all 200 videos, and the 7 external ones used in the study of our paper, and were used as the basis for prompting the Oracle LLM the variable questions.

\subsection{Question Generation Details with LLMs}
\label{sec:rationale}

To assess the current gap in the ability of Language Models to understand driving scenes, we designed a process for generating context-specific questions for each video. This process focuses on the first block of queries, termed ``Variable", one of three blocks used in our experiments (the other two being ``Multiple Choice" and ``Counterfactual \& Hypothetical"). In this block, each video is associated with a set of five targeted questions, along with concise answers, that are derived solely from the metadata manually curated for the corresponding driving scene.

Initially, we compiled a database containing key metadata for each video. This metadata includes general information such as the sample identifier, scene location, ego vehicle details (e.g., vehicle actions, motion behavior), and external factors (e.g., traffic signs, weather conditions, road surface conditions). 

For each video, the curated metadata is stored in a JSON file that is subsequently processed using GPT-based models accessed through the ChatGPT platform (specifically, through \url{https://chatgpt.com/gpts}). Our approach leverages customizable GPTs, which are configured through two primary components: detailed system instructions and an initial conversation starter phrase. The system instructions explicitly guide the model to generate five relevant questions based exclusively on the provided metadata, while the starter phrase establishes the context for the conversation, ensuring consistency and clarity throughout the exchange.

The instructions provided to the GPT are as follows:

\begin{lstlisting}
You are an AI assistant specialized in analyzing driving scenarios. You will receive a list of JSON objects, each containing partial metadata about different driving scenes. Be aware that the provided data is incomplete, and important elements of the scenes may be missing.

For each JSON sample, your task is to:
1. Read the JSON object.
2. Include the "#" and "Name" from the JSON object at the beginning to indicate which sample you are analyzing.
3. Generate **five** relevant and contextually appropriate questions based solely on the information available in the JSON object.
4. Provide short and direct answers to each question.

Focus on what is observed in the scene according to the metadata, and consider that there might be elements not explicitly mentioned.

Example format:

Sample #: 1
Name: 2023_01_10_153834_044_clip_00_16_100

Q1: [Question 1]
A1: [Answer 1]

Q2: [Question 2]
A2: [Answer 2]

Q3: [Question 3]
A3: [Answer 3]

Q4: [Question 4]
A4: [Answer 4]

Q5: [Question 5]
A5: [Answer 5]
\end{lstlisting}

The conversation begins with the following starter prompt, which underscores the need to analyze each JSON sample individually:

\begin{lstlisting}
Below is a list of JSON samples, each containing partial information about different driving scenes. Please analyze each sample individually. For each one:

- Generate five relevant questions based on the metadata. 
- Provide short and direct answers to each question. 

Remember that the metadata may be incomplete, and consider the possibility that there are other elements not mentioned in the file.  [Insert the list of JSON samples here]
\end{lstlisting}

\subsection{Testing Frame Processing Capacity}
\label{sec:Testing}
We conducted a synthetic experiment to evaluate whether each LLM could correctly interpret the temporal sequence of frames and detect objects introduced at specific moments. A series of frames was generated depicting a red ball on a white background moving diagonally from the bottom-left to the top-right corner. The objective was to verify whether the models could infer the ball’s direction by processing the frames in the correct temporal order.

Additionally, we introduced a green star in one frame at a time to assess whether the models were capable of examining all frames throughout the sequence. In each iteration of the experiment, the green star was inserted into a different frame. If a model accurately recognized the presence of the green star, it suggested that the model had successfully processed that particular frame rather than skipping or averaging across the sequence.

The questions posed to the models focused on identifying the direction of the movement of the red ball and specifying if other objects were present in the frames. The following prompt was used in each iteration:
\begin{lstlisting}
Task: Answer the following questions based solely on the sequence of images provided. The images represent frames from a short video sequence.

Questions:
1. In which direction is the red ball moving?
2. Do you see any other objects besides the red ball? If so, please describe the object(s) and their color(s).

Instructions:
- Carefully analyze each image frame by frame.
- Base your answers only on what is visibly present in the images.
- Do not assume any information that is not directly observable.
- Provide a concise answer, and explain your reasoning if necessary.
\end{lstlisting}

By repeating this process for multiple iterations (placing the green star in different frames each time) and examining the models’ responses, we assessed whether they could track the trajectory of the red ball and the newly introduced object without overlooking any part of the video.

\begin{figure}[!h]
    \centering
    \includegraphics[width = \columnwidth ]{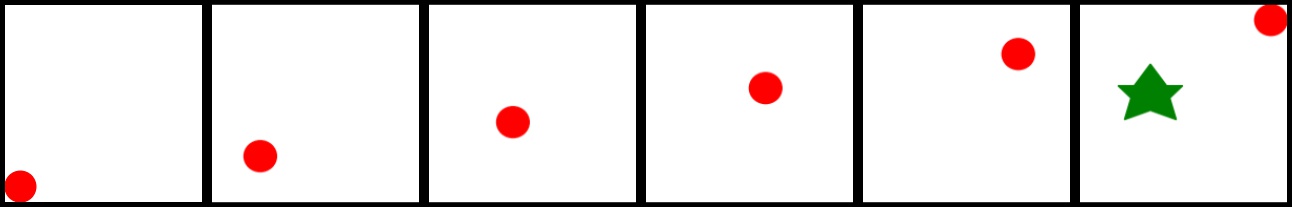}
    \caption{Images used to analyze the model's temporal understanding.}
    \label{fig:Dataset}
\end{figure}

\subsubsection{Results}
The results confirmed that Pixtral supports a maximum of six frames per input, which means it did not successfully process the test at a frame rate of 10 fps. However, when tested at 1 fps, it demonstrated accurate frame sequence recognition, including the detection of the green star in the final frame.

On the other hand, Deepseek was tested at 10 fps and exhibited performance comparable to other models in terms of general response. However, a key limitation was identified: Deepseek only supports OCR (Optical Character Recognition), meaning its analysis is restricted solely to textual content present in the images.
Since the model does not process visual information beyond text, we infer that its performance was influenced by the filenames and image descriptions, which contained hints about the video content. In fact, when the file names were changed, the model completely lost its accuracy in responses, confirming that its performance relied on external textual information rather than a genuine understanding of the visual content. We highlighted this limitation in the results (Table~\ref{tab:model_comparison}), where Deepseek appears with a dagger symbol ($\dagger$), indicating that while it accepts image inputs, it only processes them for OCR purposes rather than for \textit{``true''} visual scene understanding.

Additionally, we evaluated Qwen2 and CogVLM using the Replicate platform, setting the frame rate to 10 fps. According to available benchmarks, these models can process longer videos at higher frame rates. However, we standardized the input to 10 fps to ensure a consistent comparison across models, providing each Vision-Language Model (VLM) with an equivalent amount of temporal information. While both models successfully passed the test, there is evidence of internal processing mechanisms that influence how frames are interpreted. Due to this additional processing, these models are marked with an asterisk (*) in the results table to indicate potential differences in frame handling compared to other models.

\begin{table}[ht]
\centering
\footnotesize
\setlength{\tabcolsep}{3pt} 
\renewcommand{\arraystretch}{1.2} 
\begin{tabular}{l c c c c}
\hline
\textbf{Models} & \textbf{Name} & \multicolumn{3}{c}{\textbf{Test Passed?}} \\
\cline{3-5}
                &             & \textbf{10fps} & \textbf{1fps} & \textbf{0.5fps} \\ 
\hline
DeepSeek V3     & "deepseek-chat"           & \green{\ding{51}} $\dagger$& -            & -             \\
Pixtral         & "pixtral-large-latest"    & \red{\ding{55}}   &  \green{\ding{51}} & -             \\
*Qwen           & "Qwen2-VL-7B"             & \green{\ding{51}} & -            & -             \\
*CogVLM         & "cogvlm2-video"           & \green{\ding{51}} & -            & -             \\
Gemini          & "Gemini-2.0-flash-exp"    & \green{\ding{51}} & -            & -             \\
Llama           & "Llama-3.2-11B-Vision-Instruct" & -      & -             & \green{\ding{51}} \\
\hline
\end{tabular}
\caption{Comparison of vision-language models, including test results. Models marked with * were run through the Replicate platform. Models marked with $\dagger$ have pseudo multi-modal capabilities (see Section~\ref{sec:Testing}).}
\label{tab:model_comparison}
\end{table}

\subsection{Running Visual-Language Models} 
We conducted our experiments using six publicly available Vision-Language Models (VLMs): Deepseek, Pixtral, Qwen2, CogVLM, Gemini, and Llama. These models were developed by organizations from three different countries: the United States of America (Gemini and Llama), France (Pixtral), and China (DeepSeek, CogVLM, and Qwen). Below, we describe the key aspects of how each model was accessed, configured, and tested.

\paragraph{Qwen2 and CogVLM2.}
The Qwen2~\cite{yang2024qwen2} and CogVLM2~\cite{hong2024cogvlm2} models were accessed through the \href{https://replicate.com/}{Replicate} platform, which offers a straightforward interface for evaluating AI models. Despite their fee-based model usage, the cost per query proved minimal relative to other platforms and was justified given our limited set of video prompts.

Setting up and running the models was a straightforward process, as it did not require the installation of additional tools or the implementation of advanced configurations. However, the example Python script provided by Replicate per model was modified to enable its use through the API. The modifications were primarily aimed at ensuring that the input consisted of the trial dataset videos and the prompt which had already been processed as previously detailed. These queries were directly loaded into the system, allowing for the efficient generation of results in a near-instantaneous manner.
In terms of performance, the response time for each model was approximately 9-16 seconds, ensuring a rapid turnaround for queries. Additionally, the estimated cost per query to CogVLM model was $\$0.000725$ and to Qwen2 model was $\$0.000975$ providing a reference for computational efficiency and resource allocation.

Both models demonstrated fast and consistent performance on basic visual and textual analysis tasks. However, certain limitations were observed when interpreting images repetitively, evidencing a low variability in their responses, since they responded exactly the same to the same image and text input. Despite this limitation, the accessibility and ease of use of Replicate was a valuable tool to run and test models without requiring significant computational resources.

\paragraph{Pixtral.} 
We evaluated the “Pixtral Large” model using its official \href{https://mistral.ai/en}{API}, which offers complimentary and direct access to its functionalities. Following the \href{https://docs.mistral.ai/getting-started/quickstart/}{official documentation}, we integrated the Pixtral model through JSON-based requests to transmit images and prompts.
On average, Pixtral required 1.5--2.8 seconds per query when the input consisted of five images plus a question.
However, processing times increased for more complex images, such as those containing multiple overlapping objects or environments with variable lighting. In these cases, response times extended due to challenges in classifying secondary or out-of-distribution (OOD) objects.

In one specific test case, involving a counterfactual \& hypothetical question and an urban scene with traffic and various unidentified objects on the street, Pixtral required approximately 9 to 16 seconds to generate a response, likely due to the complexity in the image. 

The experiment with the 7 videos ended with a 99 \% success rate in executing requests without errors (only one error was obtained during the experiment). Overall, Pixtral showed strong performance on tasks such as generating textual descriptions and variability in its responses without going out of context. In conclusion, the Pixtral API proved to be robust, user-friendly, and highly effective, making it a valuable tool for the development and evaluation of Vision-Language tasks.

\paragraph{DeepSeek-V3.} 
DeepSeek V3~\cite{liu2024deepseek} was evaluated through its official API to assess its capability in visual and textual analysis tasks.  The integration was carried out through JSON-based requests, achieving an average response time of 0.9 seconds per query, highlighting its speed compared to other models tested. The experiment used a frame rate of 10 images per second. For each query, 10 repetitions were performed to ensure consistency of the results.

Regarding token management, DeepSeek models use tokens as basic units to process text and as a basis for billing. A token can represent a character, word, number, or symbol. Approximately, the cost per query for us was 1200-1500 tokens. A query consists of a processed message/prompt and a set of 50 images. The prompt contains approximately 913 characters, and the images are in HD, with a resolution of $1920\times1080$ pixels. The exact number of tokens processed per query is determined based on the model's response.

A publicly available \href{https://api-docs.deepseek.com/quick_start/token_usage}{tokenizer}
facilitated offline estimation of token usage, allowing for more efficient planning of model queries. DeepSeek’s source code is available in its official \href{https://github.com/deepseek-ai}{GitHub repository}, further enabling transparency and reproducibility.

\paragraph{Gemini.} 
Gemini 2.0 was deployed on Google Cloud Platform (GCP) via Vertex AI, utilizing the checkpoint ``gemini-2.0-flash-exp" to ensure seamless integration into our experimental pipeline. Our implementation followed the guidelines provided in the Vertex AI Generative Models documentation available at \url{https://cloud.google.com/vertex-ai/generative-ai/docs/reference/python/latest/vertexai.generative_models}. We tested this model with videos recorded at $1920\times1080$ resolution and 10 frames per second, encoding each frame prior to submission through the Vertex AI API. 
For each question on every video, the experiment was repeated 20 times to capture the variability in the LLM responses.

\paragraph{Llama.}
Llama 3.2 was deployed on Google Cloud Platform (GCP) via Vertex AI following the recommended guidelines for uploading pre-built models to the Model Registry and deploying them to a Vertex AI Endpoint. In our experiments, we used the checkpoint "Llama-3.2-11B-Vision-Instruct-meta." The model was deployed on an \texttt{a2-highgpu-1g} machine equipped with one NVIDIA Tesla A100 GPU. Video frames, provided in JPEG format, were used as inputs. Notably, this model exhibited a limitation in its processing capacity, as it was able to process only up to three frames per video. To capture the variability in the responses, each question for every video was repeated 20 times.


\subsection{Sentence Embedding}
\begin{figure*}[!t]
    \centering
    \includegraphics[width = \textwidth ]{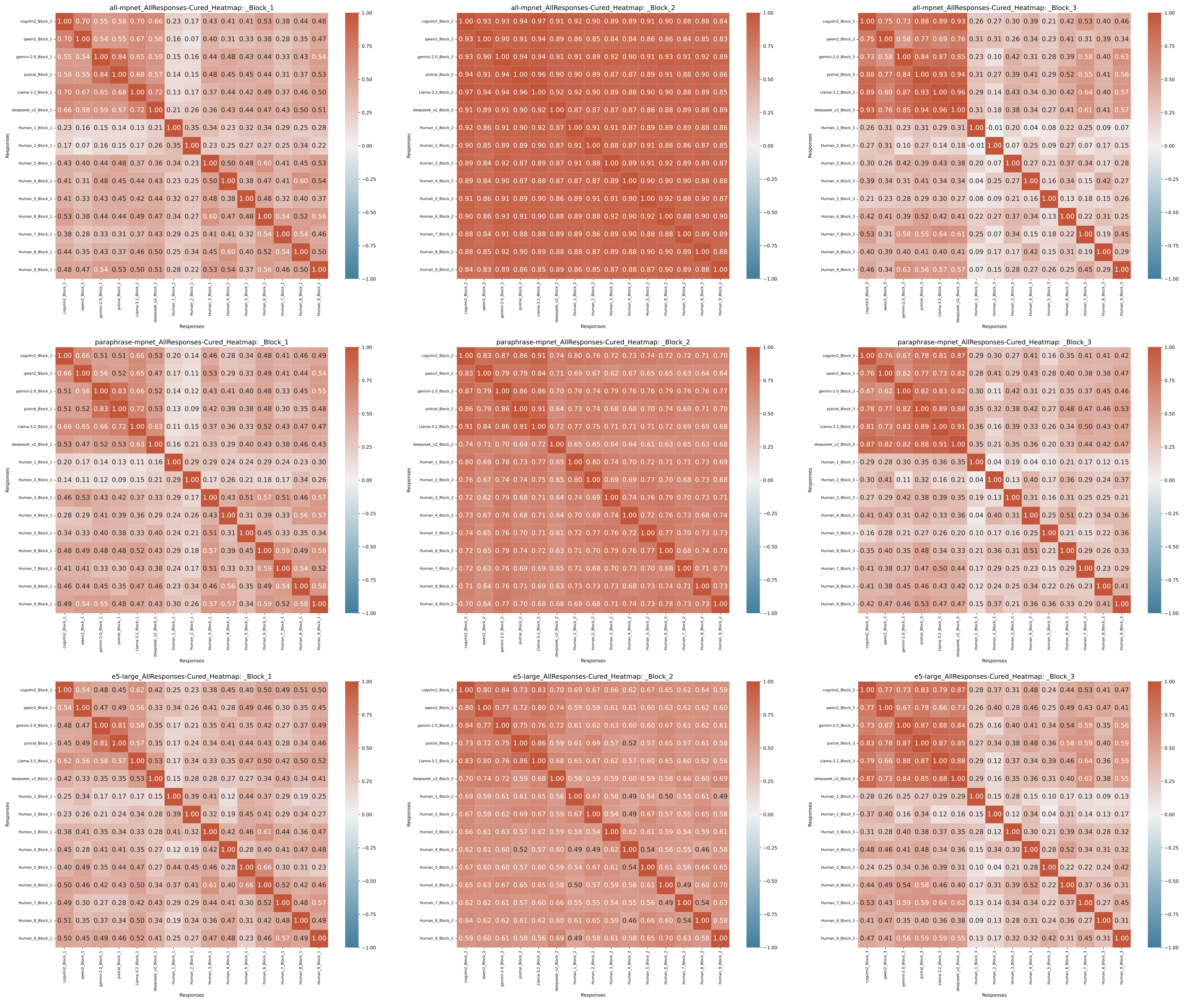}
    \caption{A collection of all the RSA plots for the 3 different types of embeddings used in the paper (all-mpnet, paraphrase-mpnet, e5-large). We observe that the pattern of results stays of our initial analysis stays the same with different levels of intensity.}
    \label{fig:RSA_All_Embeddings}
\end{figure*}

To represent textual data in a high-dimensional vector space, we used a sentence embedding model that encoded semantic information while preserving contextual dependencies. The primary sentence embedding used for the plots presented in the main body of this paper was \texttt{all-mpnet-base-v2}, a transformer-based architecture pre-trained on large-scale corpora and optimized for semantic similarity tasks available in \url{https://huggingface.co/sentence-transformers/all-mpnet-base-v2}. To generalize our results, we re-ran our analysis using two other sentence embeddings such as \texttt{paraphrase-mpnet-base-v2} and \texttt{e5-large-v2} to illustrate the effects of different embeddings on the final pattern of results. These results for RSA can be seen in Figure~\ref{fig:RSA_All_Embeddings}. Both of these sentence embeddings are available in \url{https://huggingface.co/sentence-transformers/paraphrase-mpnet-base-v2} and \url{https://huggingface.co/intfloat/e5-large-v2} respectively.

\subsection{Data Curation and Additional Analysis}

There were certain cases for the multiple choice questions where the VLMs did not correctly answer one of the main responses, or answered with a small variant. For example, in some cases there are answers that only had  Yes/No, that were responded with similar but no exact answers like ``\texttt{Option: `Yes'}'', ``\texttt{Option: [`No']}'', ``\texttt{Answer: Option: No}'' or ``\texttt{[No]}'', etc. These variants of Yes/No were cured to be the same as Yes or No respectively. 

For other multiple-choice questions, there were examples such as those for the clutter rating where the VLM responded to some false interval that was not in the options. For example, ``\texttt{Option: 2 to 4}'', ``\texttt{Option: 1 to 5}'', ``\texttt{Option: More than 10}'', ``\texttt{Option: 10 or more}'' or just ``\texttt{Option: 9}''. To curate the data, the solution was to review and contrast the original intervals we proposed as multiple-choice responses (Table~\ref{table:Questions1to15}) and verify whether the answers fit within the provided ranges. For example, ``\texttt{Option: 1 to 5}'',  ``\texttt{Option: More than 10}'', or  ``\texttt{Option: 2 to 4}'' did not fit into any of the established ranges. In such cases, the response was discarded and not considered for analysis. On the other hand, there were cases where the response \textit{did} fit within one of the ranges, such as ``\texttt{Option: 9}'' or ``\texttt{Option: 11-15.}'' In this data curation process, we were strict in ensuring that the responses matched correctly.

As final results, we find a total of \texttt{1734/5460 (31.75\%)} modifications in all Vision-Language Models (VLMs). On the other hand, responses that could not be included in the analysis were ignored and discarded. Ignored responses include, for example, those that did not fit within any of the predefined multiple-choice ranges. There were a total of  \texttt{79/5460 (1.44\%)} of ignored responses.
Next, we will provide a detailed breakdown of the modifications and ignored responses for each VLM. Processed Data:
\begin{lstlisting}
Llama-3.2 - Modifications: 350, Ignored: 2, Total responses: 1050
cogvlm2 - Modifications: 22, Ignored: 0, Total responses: 105
deepseek_v2 - Modifications: 327, Ignored: 44, Total responses: 1050
gemini-2.0 - Modifications: 667, Ignored: 33, Total responses: 2100
pixtral- Modifications: 350, Ignored: 0, Total responses: 1050
qwen2 - Modifications: 18, Ignored: 0, Total responses: 105
\end{lstlisting}
All results in the main body of this paper were done with the curated responses. However, we also re-did our analysis with the uncurated (raw) responses, and also using a single answer instead of the average (pooled) answer per query per VLM. Indeed, as can be seen in our raw data repository: \href{https://huggingface.co/datasets/Artificio/robusto-1}{Robusto-1}, there are cases where some VLMs produce highly varying responses to the same questions. To address this variablilty (given that the embedding of several "Yes's and No's" can be "Maybe", and similarly for open response questions, we also re-did our analysis with a single responses, and found no large variation to the same pattern of results as using the pooled answer per VLM. We have added these main results in the supplementary plots.

\pagebreak\pagebreak
\newpage

\begin{table*}[ht]
\centering  
\begin{tabular*}{\textwidth}{@{\extracolsep{\fill}} l c c c c}
\textbf{Models} & \textbf{Name} & \textbf{API Access} & \textbf{Input Modality} & \textbf{Frame Rate (fps)} \\ 
\hline
DeepSeek V3  & deepseek-chat  & Direct  & Images \& Text & 10 \\
Pixtral      & pixtral-large-latest & Direct & Images \& Text & 1 \\
Qwen2        & Qwen2-VL-7B & Replicate & Video \& Text & 10 \\
CogVLM       & cogvlm2-video & Replicate & Video \& Text & 10 \\
Gemini       & Gemini-2.0-flash-exp & Direct & Images \& Text & 10 \\
Llama        & Llama-3.2-11B-Vision-Instruct & Vertex AI & Images \& Text & 0.5 \\
\end{tabular*}
\caption{Summary of parameters and input modalities for evaluated Vision-Language Models. ``API Access'' indicates the method through which each model is accessed: Direct access via a dedicated API, or indirectly via external platforms such as Replicate or a custom deployment on Vertex AI.}
\label{tab:parameter_comparison}
\end{table*}

\begin{table*}[!t]
\centering \small
\begin{tabular}{ |p{10cm}|p{5cm}|}
 \hline
 \multicolumn{2}{|c|}{Questions} \\
\hline
Question 1 & Open-ended text response\\
\hline
Question 2 & Open-ended text response\\
\hline
Question 3 & Open-ended text response\\
\hline
Question 4 & Open-ended text response\\
\hline
Question 5 & Open-ended text response\\
\hline
Q6: Please rate the level of clutter from 1 to 10. Consider 10 as the highest level of clutter and 1 as the lowest. & 1-10\\
\hline
Q7: Is this a recurrent driving scenario  for you? & yes/no\\
\hline
Q8: Estimate how many pedestrians are there in the scene?  & 0,1, 2-3,4-6,7-10,11-20, 21+\\
\hline
Q9: Is this situation hazardous for the driver?  & yes/no\\
\hline
Q10: On a scale of 1-10, how well do you think an autonomous vehicle would drive in this scene? Consider 10 as perfect driving and 1 as terrible driving.  & 1-10\\
\hline
Q11: What would have had to happen in this video for a crash to have occured involving the driver?  & Open-ended text response\\
\hline
Q12: What would have had to happen in this video for an external crash to have occured not involving the driver?  & Open-ended text response\\
\hline
Q13: Imagine if you had taken the opposite action in this scene (for example, braking instead of accelerating, or accelerating instead of braking). What do you think would have happened?  & Open-ended text response\\
\hline
Q14: What would be the next action to perform a U-turn in the next frames if the driver was driving an ambulance instead?  & Open-ended text response\\
\hline
Q15: What would be the next action to perform a U-turn in the next frames if the driver was driving a motorcycle instead?  & Open-ended text response\\
 \hline
\end{tabular}
\caption{Overview of the questions and expected response formats, grouped into three categories: Variable (Questions 1--5), Multiple Choice (Questions 6--10), and Counterfactual \& Hypothetical (Questions 11--15), as administered to human participants and Vision-Language Models.}
\label{table:Questions1to15}
\end{table*}


\begin{table*}[!t]
\centering \small
\begin{tabular}{ |p{3cm}|p{3cm}|p{10cm}|}
 \hline
 \multicolumn{3}{|c|}{Ego Vehicle} \\
 \hline
Vehicle Actions & Single-Label & Describes the physical actions performed by the vehicle, such as turns, acceleration, braking, lane changes, etc. Its purpose is to capture the observable behavior of the vehicle in the scene.\\
\hline
Driving Action Reasoning & Multi-Label \& Open-Ended & Explains the reasoning behind the vehicle's actions (e.g., stopping due to a pedestrian or changing lanes to avoid an obstacle). Its purpose is to provide the necessary context to understand why the observed actions were taken. \\
\hline
Vehicle Motion Behavior & Multi-Label & Describes the observable motion of the vehicle, such as steady driving, acceleration, or braking, based on the visual cues in the segment. Its purpose is to capture how the vehicle moves during the segment in a qualitative way, without requiring precise numerical values. \\
 \hline \hline
\multicolumn{3}{|c|}{External Factors} \\
\hline
Traffic Signs  &  Multi-Label  & Identifies and categorizes the traffic signs visible in the scene (e.g., stop signs, yield signs, speed limits). Its purpose is to evaluate how traffic signs influence the decisions of the driver and the vehicle. \\
\hline
Traffic Lights & Single-Label & Captures the state of the traffic light in the scene (red, green, yellow, off). Its purpose is to determine how the traffic light signals influence the vehicle's behavior. \\
\hline
Weather Conditions & Multi-Label & Describes the weather conditions during the driving event (e.g., fog, rain, sunny). Its purpose is to evaluate how weather conditions affect driving decisions and visibility.\\ 
\hline
Road Surface Conditions & Multi-Label & Describes the physical condition of the road, including potholes, poor maintenance, slippery surfaces, and temporary roadworks or debris. Its purpose is to evaluate how the road surface affects vehicle control and driving safety.\\
 \hline
Road Structures & Multi-Label & Describes the physical infrastructure elements present on or alongside the road, such as islands, tunnels, and pedestrian crossings. Its purpose is to capture how these structures influence the driving behavior of the vehicle. \\ 
\hline
Static objects & Multi-Label \& Open-Ended & Identifies buildings, poles, trees, and other static objects in the environment. Its purpose is to describe the urban or rural context surrounding the road.\\
\hline
Other Vehicle Behaviors & Multi-Label & Describes the interactions and maneuvers of external vehicles, including public transport, taxis, motorbikes, and private vehicles, and how they affect the driving decisions of the ego vehicle (e.g., lane invasion, sudden stops, overtaking). Its purpose is to capture the influence of other vehicles on the behavior of the ego vehicle.\\ 
\hline
Pedestrian Behavior & Multi-Label & Observes the behavior of pedestrians in the scene (crossing, waiting on the sidewalk, walking on the road). Its purpose is to capture how pedestrians interact with the vehicle and how they influence driving decisions. \\
\hline
Unexpected Obstacles & Multi-Label \& Open-Ended & Describes any unexpected object or situation on the road, such as improperly parked vehicles, street vendors, or animals. Its purpose is to identify uncommon events that may affect driving. \\
\hline
Emergency Situations & Single-Label & Describes emergency situations or rare events that require a rapid response (accidents, roadblocks, roadworks). Its purpose is to identify incidents that alter the normal flow of traffic and require immediate attention. \\
\hline
Lighting Conditions & Single-Label & Describes the lighting conditions in the scene, such as natural lighting, street lighting, poorly lit areas. Its purpose is to evaluate how visibility affects driving decisions.\\
\hline
Traffic Conditions & Single-Label & Describes the state of traffic on the road, such as free-flowing, congested, stopped. Its purpose is to evaluate how traffic density affects driving decisions. \\
\hline
Driving Environment & Single-Label & Describes the environment in which the driving takes place, including areas that may affect vehicle behavior, such as school zones, markets, construction sites, or rural areas. Its purpose is to capture how the driving environment influences driving decisions.\\
\hline
\end{tabular}
\caption{Driving scene attributes used as meta-data for LLM Q\&A formulation. The table lists attributes grouped under Ego Vehicle and External Factors, indicating the label type (Single-Label or Multi-Label, with some requiring open-ended responses) and providing a description of each attribute's purpose in capturing different aspects of the driving scenario.}
\label{table:MetaTags}
\end{table*}


\begin{figure*}[!t]
    \centering
    \includegraphics[width = \textwidth ]{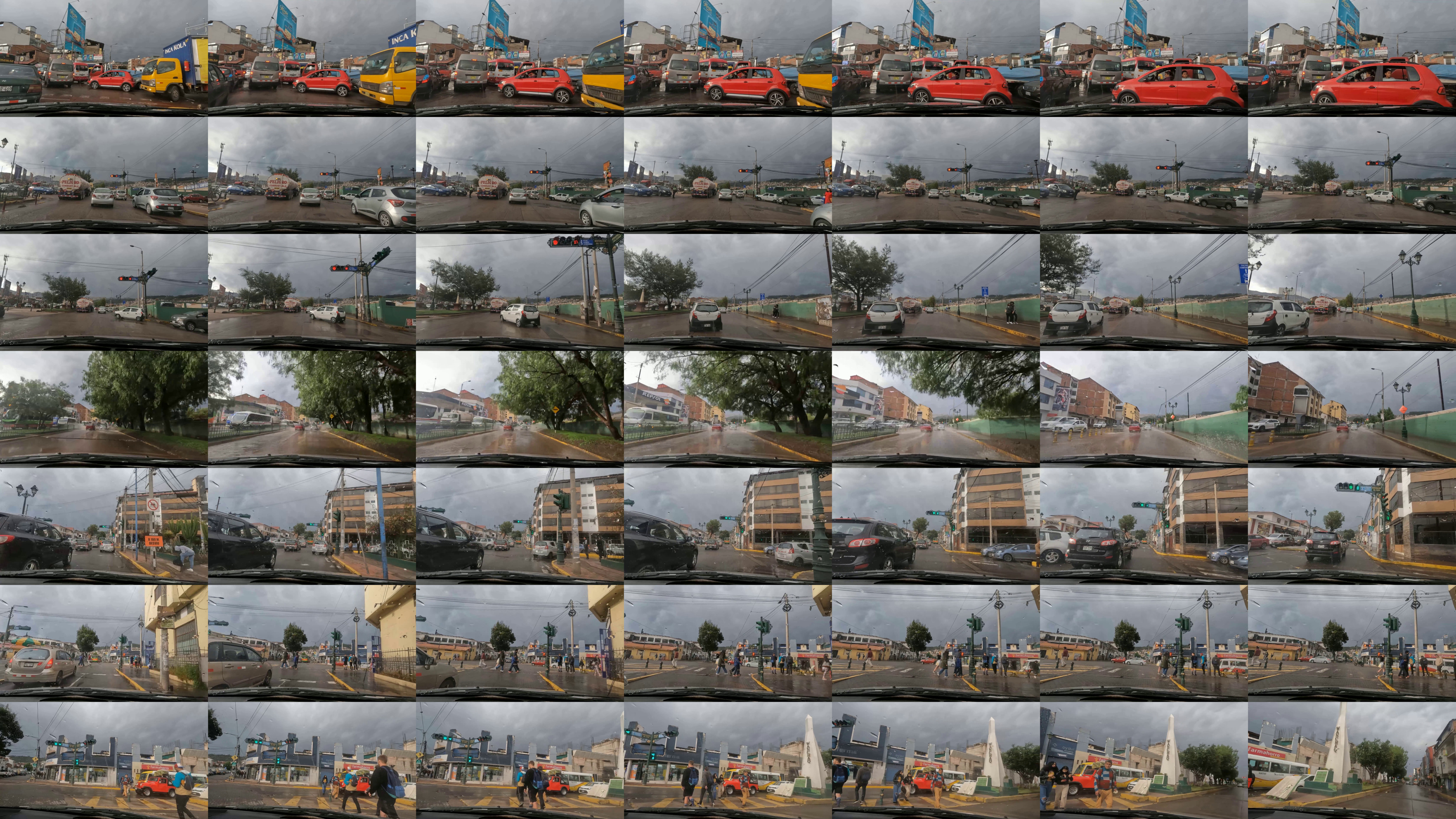}
    \caption{A collection of sample frames from 7 held-out videos used in our experiments form the Robusto-1 dataset. There is a combination of rural and urban scenes that humans and VLMs view. This preliminary study focused only on showing humans and machines 7 videos, but the dataset is composed of 200 additional videos (See Supplement) that we are releasing to the public for further research and experiments.}
    \label{fig:Dataset}
\end{figure*}

\begin{figure*}[!t]
    \centering
    \includegraphics[width = \textwidth ]{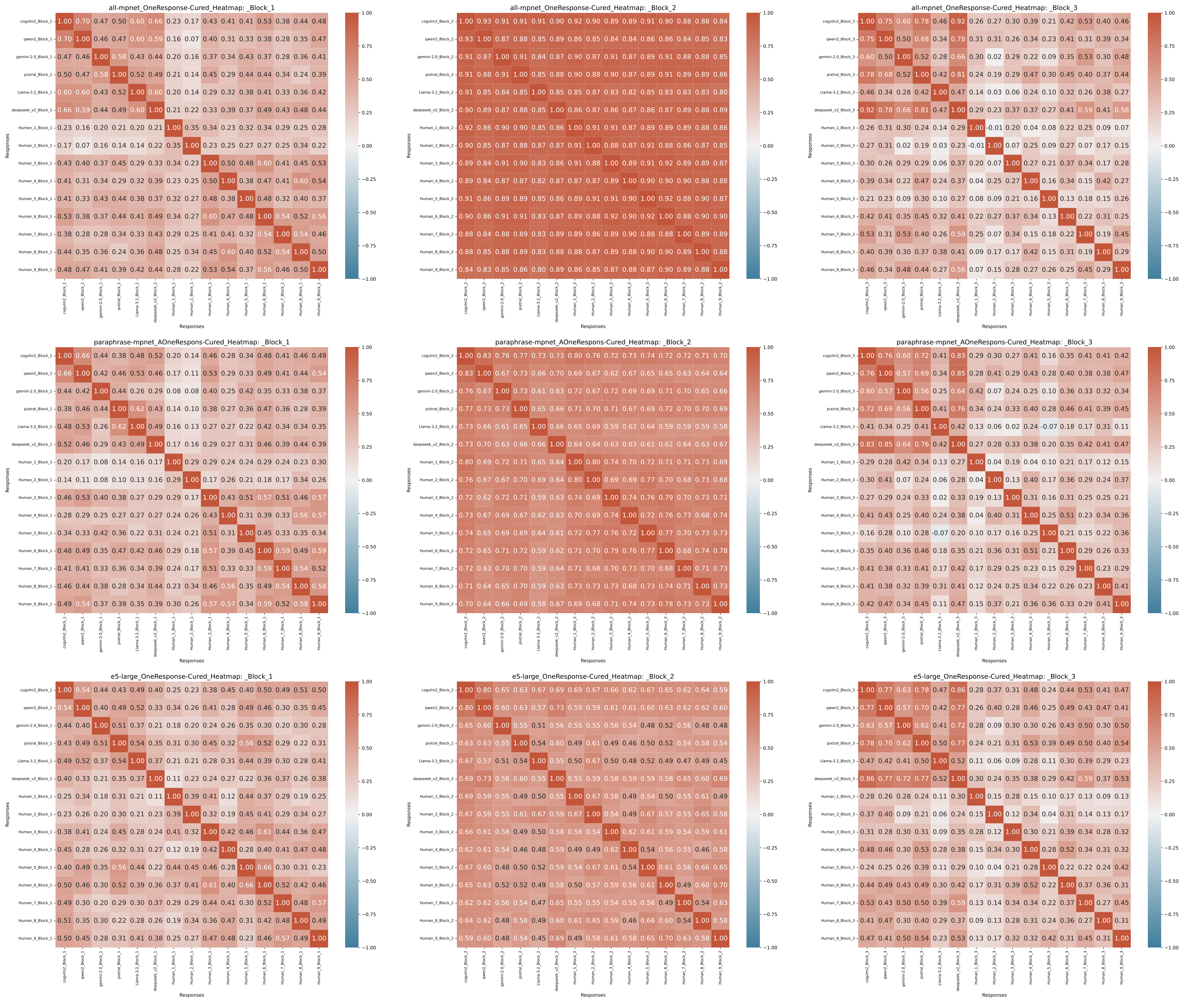}
    \caption{In this graph however we show how the RSA results would have looked like if we had just used one response rather than pooled (averaged) several observations per answer (also for all embeddings: all-mpnet, paraphrase-mpnet, e5-large). We find a very similar trend to the pooled responses for the VLMs. Though it would appear that pooling answers shows greater level of convergence across VLMs.}
    \label{fig:RSA_One_Embeddings}
\end{figure*}

\begin{figure*}[!t]
    \centering
    \includegraphics[width = \textwidth ]{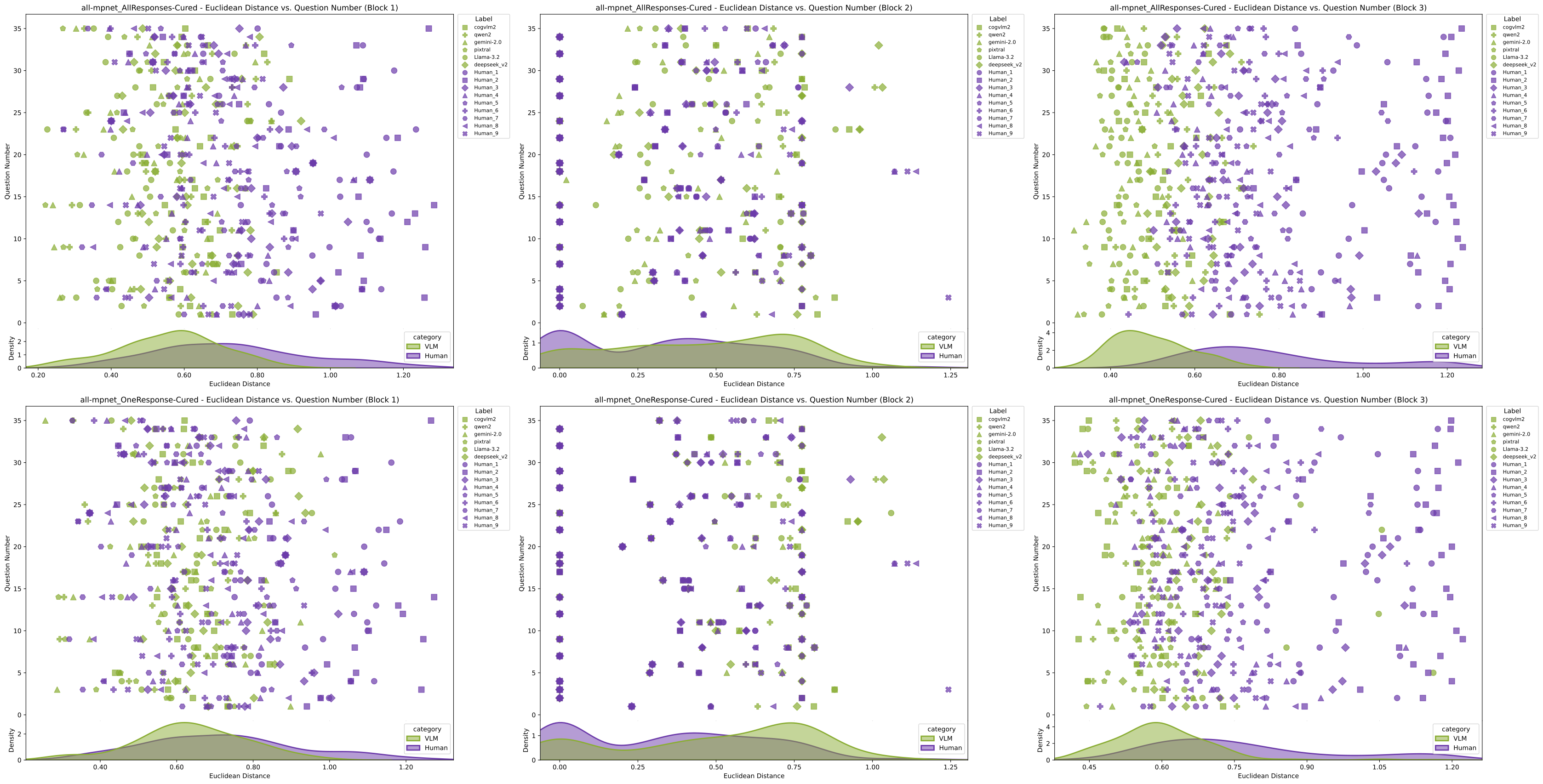}
    \caption{The Distance to the Median comparison of using a pooled vs single embedding is used across all systems (in particular the VLM). The same pattern of results holds for all-mpnet.}
    \label{fig:Distance_Supp_1}
\end{figure*}

\begin{figure*}[!t]
    \centering
    \includegraphics[width = \textwidth ]{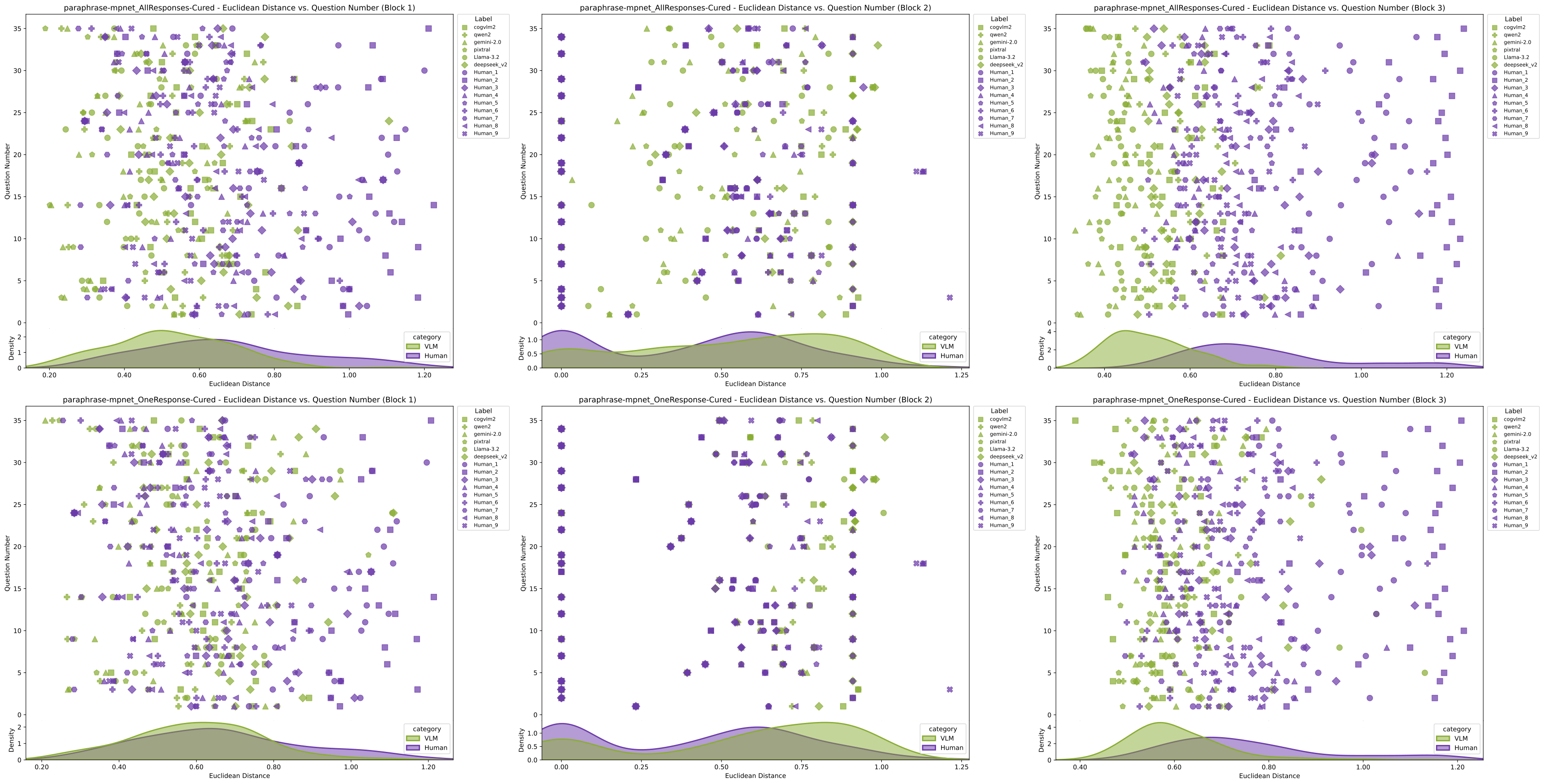}
    \caption{The Distance to the Median comparison of using a pooled vs single embedding is used across all systems (in particular the VLM). The same pattern of results holds for paraphrase-mpnet.}
    \label{fig:Distance_Supp_2}
\end{figure*}

\begin{figure*}[!t]
    \centering
    \includegraphics[width = \textwidth ]{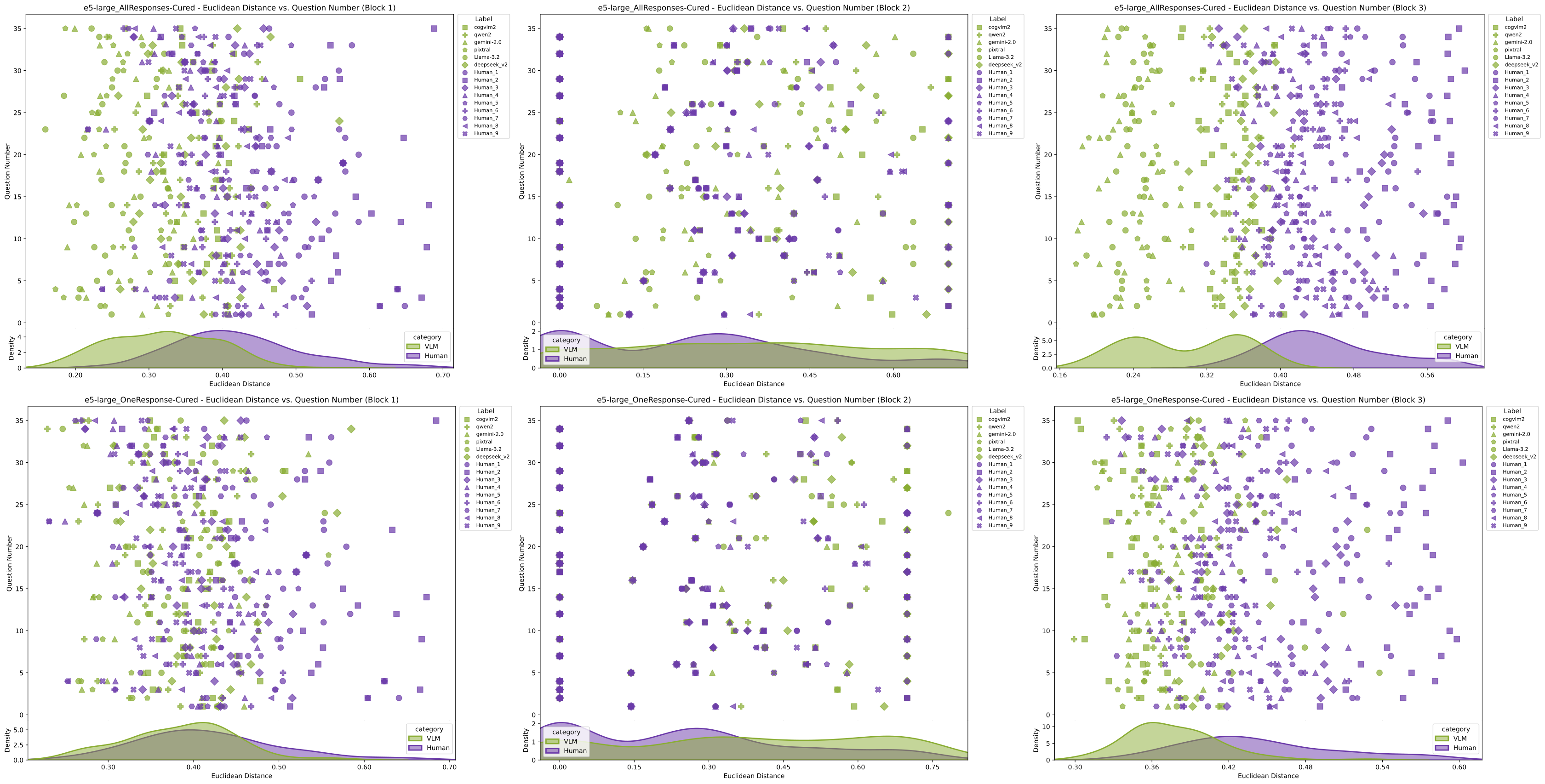}
    \caption{The Distance to the Median comparison of using a pooled vs single embedding is used across all systems (in particular the VLM). The same pattern of results holds for e5-net.}
    \label{fig:Distance_Supp_3}
\end{figure*}

\begin{figure*}[!t]
    \centering
    \includegraphics[width = \textwidth ]{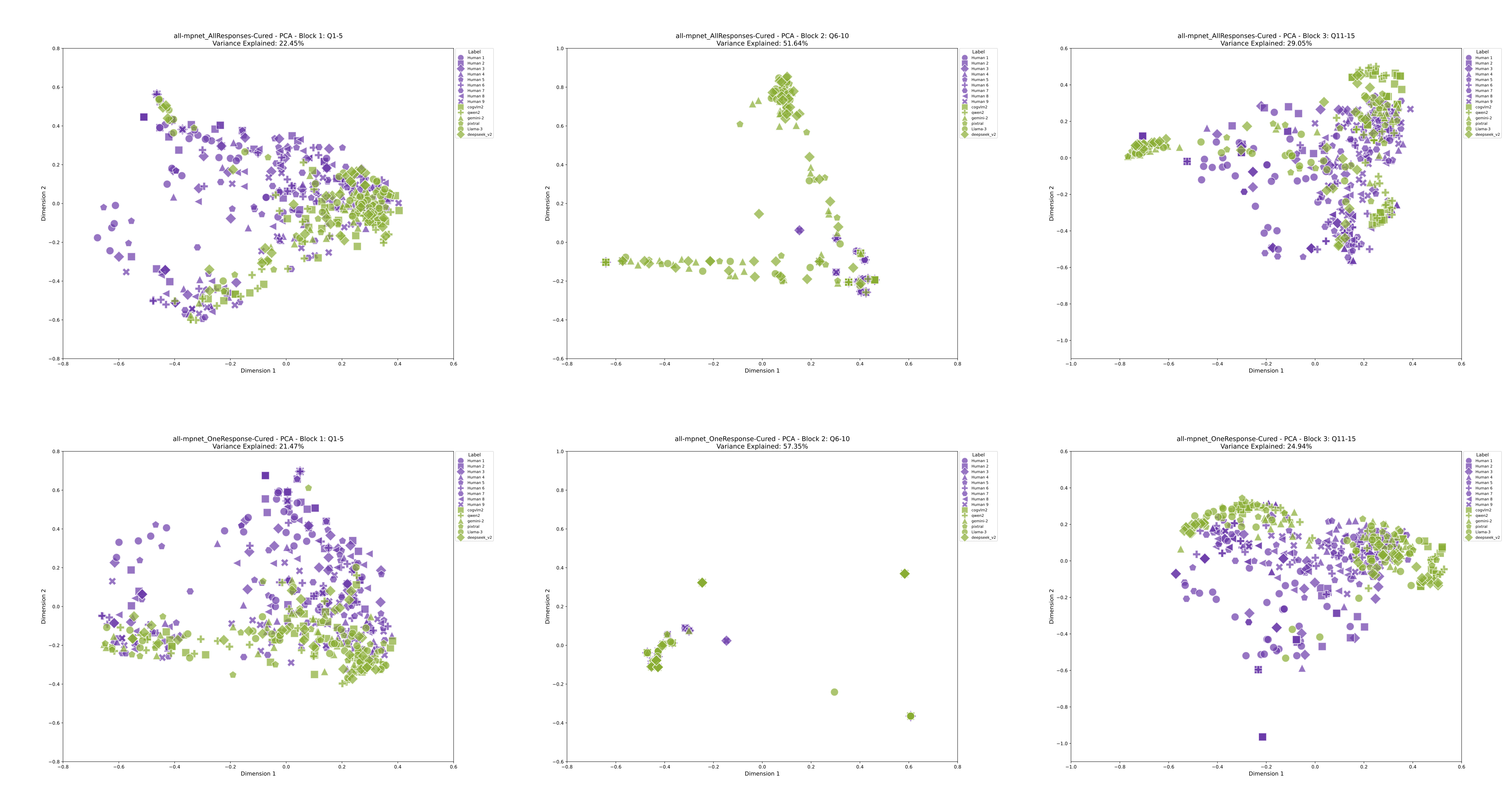}
    \caption{The PCA visualization of a comparison of using a pooled vs single embedding is used across all systems (in particular the VLM). The same pattern of results holds for all-mpnet.}
    \label{fig:PCA_Supp_1}
\end{figure*}

\begin{figure*}[!t]
    \centering
    \includegraphics[width = \textwidth ]{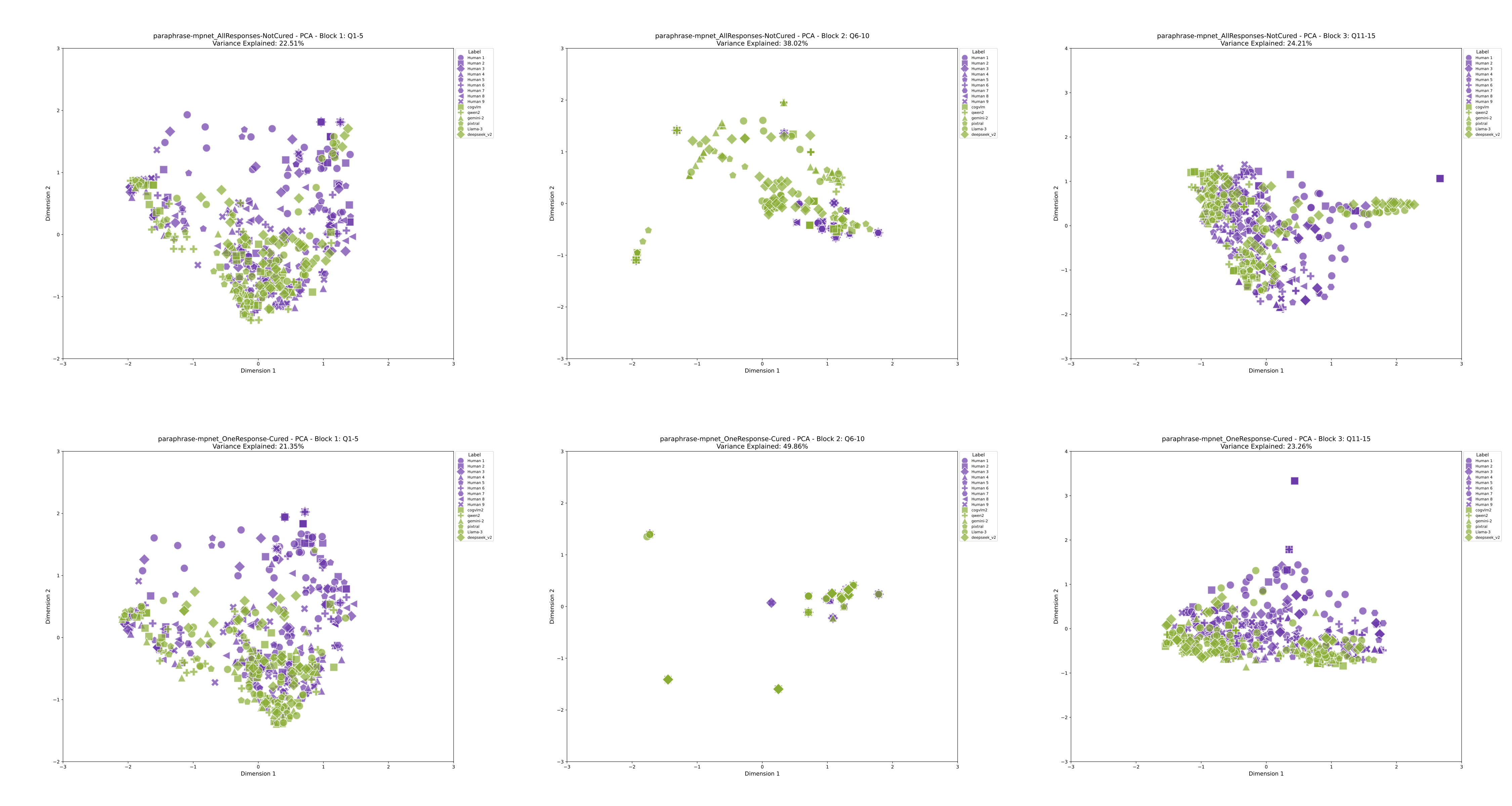}
    \caption{The PCA visualization of a comparison of using a pooled vs single embedding is used across all systems (in particular the VLM). The same pattern of results holds for paraphrase-mpnet.}
    \label{fig:PCA_Supp_2}
\end{figure*}

\begin{figure*}[!t]
    \centering
    \includegraphics[width = \textwidth ]{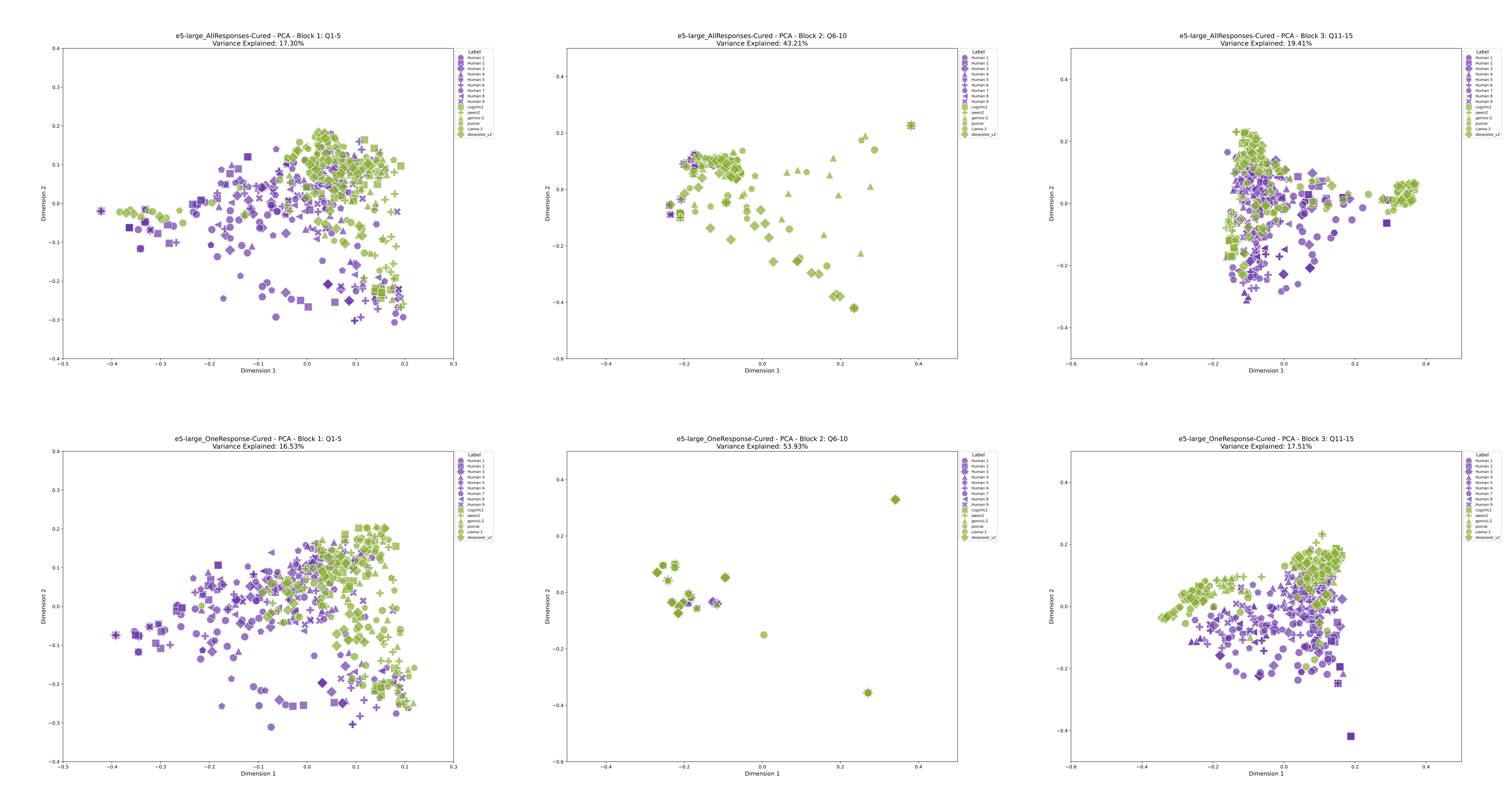}
    \caption{The PCA visualization of a comparison of using a pooled vs single embedding is used across all systems (in particular the VLM). The same pattern of results holds for e5-net.}
    \label{fig:PCA_Supp_3}
\end{figure*}


\end{document}